%% file: XMemTransfer-main.tex
\documentclass{article}
\usepackage[T1]{fontenc}
\usepackage[margin=2.8cm]{geometry}
\usepackage{graphicx}
\usepackage{picinpar}
\usepackage{subcaption}
\usepackage{xcolor}
\usepackage{colortbl}
\usepackage{booktabs}
\usepackage{multirow}
\usepackage{soul}
\usepackage{enumerate}

\usepackage{tikz}
\usetikzlibrary{arrows.meta, backgrounds}
\usepackage{amsmath}
\usepackage{amssymb}
\usepackage{pifont}
\usepackage{tabularx}
\usepackage{array}
\usepackage{makecell}
\usepackage[ruled,vlined,linesnumbered]{algorithm2e}

\usepackage{threeparttable}

\usepackage{hyperref} 
\usepackage{cleveref}            
\AddToHook{cmd/appendix/before}{%
    \crefalias{section}{appendix}%
    \crefalias{subsection}{appendix}%
    \crefalias{subappendix}{appendix}%
    \crefalias{subsubappendix}{appendix}
}
\usepackage{kpfonts}             
\usepackage{natbib}              
\usepackage{tcolorbox}           
\usepackage{fontawesome5}        
\usepackage{xspace}              
\usepackage{microtype}           

\usepackage[]{authblk}
\usepackage{fancyhdr}
\fancypagestyle{firstpage}{%
  \fancyhead[L]{\raisebox{0.05\height}{{\hypersetup{hidelinks}\href{https://www.olaresearch.org/}{\includegraphics[height=1.2em]{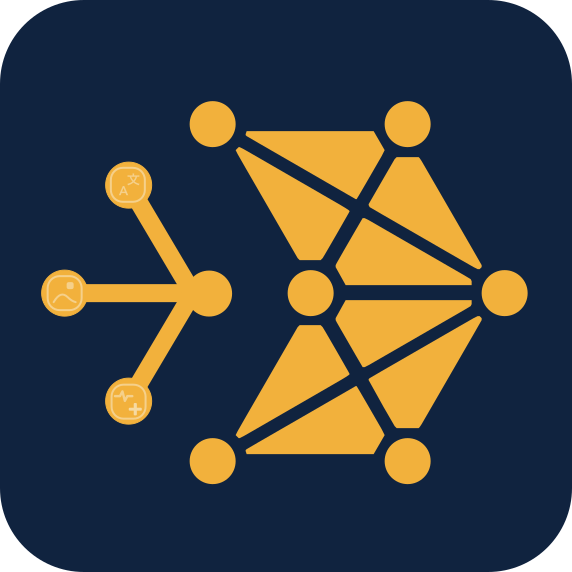}}}}~{Accepted at NeurIPS 2026}}%
  \fancyhead[R]{\raisebox{0.05\height}{{\hypersetup{hidelinks}\href{https://www.ellisinstitute.fi}{\usebox{\ellislogo}}}}}%
}

\newsavebox{\ellislogo}
\sbox{\ellislogo}{\includegraphics[height=1em]{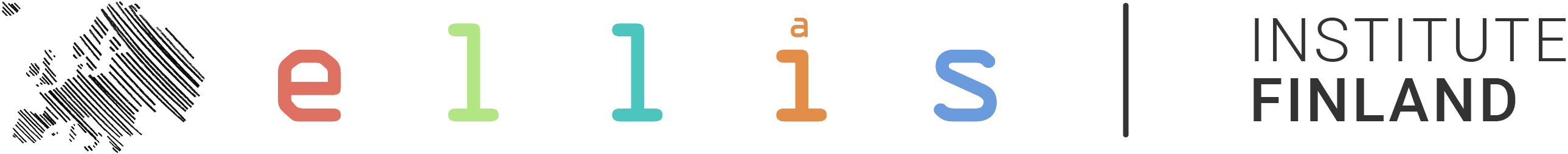}}

\setcitestyle{round, elide, numbers}
\definecolor{colornavy}{HTML}{1B365D}      
\definecolor{coloramber}{HTML}{E0913A}     
\definecolor{colorviolet}{HTML}{8B5DCB}    
\definecolor{colormint}{HTML}{004A43}      
\definecolor{colorblue}{HTML}{1D70B8}      
\definecolor{colorforest}{HTML}{194D33}    
\definecolor{colorsage}{HTML}{5F8575}      
\definecolor{colorburgundy}{HTML}{800020}   
\definecolor{colorcrimson}{HTML}{B30D26}   
\definecolor{colorcoral}{HTML}{E07A5F}     
\definecolor{colorindigo}{HTML}{3F51B5}    
\definecolor{colorplum}{HTML}{5C2D50}      
\definecolor{colorslate}{HTML}{475569}     

\newcommand{\furl}[1]{\footnote{\url{http://#1}}}

\definecolor{groupgray}{RGB}{246,246,246}
\definecolor{harmonyblue}{RGB}{222,244,248}
\definecolor{posgreen}{RGB}{0,130,55}
\definecolor{bestgreen}{RGB}{0,130,55}
\definecolor{negred}{RGB}{230,0,0}
\definecolor{algblue}{RGB}{40,96,160}
\definecolor{algred}{RGB}{185,55,55}

\newcommand{\pos}[1]{\textcolor{posgreen}{#1}}
\newcommand{\negc}[1]{\textcolor{negred}{#1}}
\newcommand{\cmark}{\ding{51}}
\newcommand{\xmark}{\ding{55}}
\newcommand{\pmark}{{\color{gray}$\circ$}}

\newcommand{\takeawaybox}[1]{%
    \par\smallskip \noindent \begingroup
    \setlength{\fboxsep}{6pt}%
    \colorbox{harmonyblue!55}{\parbox{\dimexpr\linewidth-2\fboxsep\relax}{\textbf{Takeaway.} #1
        }%
    }%
    \endgroup \par\smallskip
}

\usepackage{placeins}
\DeclareMathOperator{\RMSNorm}{RMSNorm}
\renewenvironment{abstract}{%
  \begin{tcolorbox}[
    colback=colornavy!4,
    colframe=colornavy,
    leftrule=4pt,
    rightrule=0.5pt,
    toprule=0.5pt,
    bottomrule=0.5pt,
    arc=3pt,
    boxsep=5pt,
    left=12pt,
    right=12pt,
    top=10pt,
    bottom=10pt,
    title=\textbf{Abstract},
    coltitle=colormint,
    attach title to upper, after title={\par\medskip}
  ]%
}{\medskip{\color{colornavy}\hrule height 0.5pt}\medskip \small\noindent
  {\hypersetup{hidelinks}%
  \begin{tabular}{@{}l@{\hspace{0.5em}}l@{}}
    \textcolor{colormint}{\faGlobe} & \textbf{Project Website:} \url{https://olaresearch.org/XMemTransfer} \\
    \textcolor{colormint}{\faGithub} & \textbf{GitHub Code:} \url{https://github.com/OLAResearch/XMemTransfer} \\
    \textcolor{colormint}{\faRobot} & \textbf{HuggingFace Models:} \url{https://hf.co/collections/OLAResearchX/xmemtransfer}
  \end{tabular}}
  \end{tcolorbox}%
}

\title{Frozen Memory Is Not Enough: Rethinking External Memory as Extraction}
\author[$\star$,$\dagger$]{Mingyuan Li}
\author[$\ddagger$]{Guangsheng Yu}
\author[$\ddagger$]{Xu Wang}
\author[$\star$,$\dagger$]{Shaoxiong Ji\thanks{Corresponding author.}}
\affil[$\star$]{ELLIS Institute Finland}
\affil[$\dagger$]{University of Turku}
\affil[$\ddagger$]{University of Technology Sydney}
\affil[]{\textit{Emails:~mingyl@utu.fi,~guangsheng.yu@uts.edu.au,~xu.wang@uts.edu.au,~shaoxiong.ji@utu.fi}}

\date{}

\begin{document}

\maketitle
\thispagestyle{firstpage}

\begin{abstract}
Methods for improving knowledge use in large language models typically fall into two regimes.
Non-parametric retrieval offers flexible access to external knowledge, but adds retrieval latency, context overhead, and only shallow integration with the backbone.
Parametric adaptation is efficient at inference time, but entangles knowledge with model weights and can be hard to update, audit, or transfer.
Engram-style hashed memory occupies a middle regime: it stores learned information in an external, addressable table, yet consumes that table through a small learned reader.
This raises a basic question: when such a memory is moved across backbones, what matters more, the frozen memory itself or the target-side reader?
We study this question through cross-model frozen-memory extraction, in which a memory trained on a source model is frozen and attached to a different target model, with only a lightweight reader trained.
Ablations show that learned memory content and correct addressing both matter, while the amount of usable signal depends on how well the reader is aligned to the target model.
In downstream question answering tasks, a dual-layer, four-branch reader nearly closes the gap between same-model and cross-model reuse, achieving an average score of 38.8 under our controlled evaluation protocol.
Moreover, when the provider reader is directly compatible with the target interface, the frozen artifact can provide substantial utility without target-side training, while optional reader adaptation yields further improvement.
These results suggest that Engram can serve as a reusable external knowledge artifact, provided that the target has access to a compatible reader interface; target-side adaptation can further improve alignment when direct reader reuse is insufficient.
\end{abstract}

\section{Introduction}
\label{sec:intro}
Language models can use knowledge in two familiar ways.
\emph{Non-parametric} methods, such as retrieval-augmented generation (RAG)~\citep{lewis2020rag} and nearest-neighbor retrieval~\citep{khandelwal2021generalization}, keep knowledge outside the backbone and make updates relatively direct, but they pay for that flexibility with retrieval latency, longer contexts, and loose integration between retrieved evidence and generation.
\emph{Parametric} methods, such as continued pretraining, fine-tuning, or learned memory layers tied to a single model~\citep{chen2025mlpmemory}, avoid retrieval at inference time, but they entangle knowledge with model weights.
This makes updates, audits, and transfers across backbones difficult.

Engram-style hashed memory~\citep{cheng2026engram} occupies an interesting middle ground.
It stores learned information in an explicit, addressable external table, yet consumes that table through a small neural interface rather than by inserting raw documents into the context.
This hybrid design is attractive because it combines the modularity of external memory with the efficiency of learned representations.
But it also exposes a sharper question: \textbf{is the memory itself reusable or transferable, or is it merely a co-adapted extension of the model that trained it?}

The distinction matters: a memory table that is not portable is just another form of model-specific parameterization, not a reusable knowledge substrate.
Thus, the portability requires an operational test: \emph{remove the source backbone and ask whether another model can still extract a useful signal from the frozen memory table.}

We study this test through \emph{cross-model frozen-memory transfer}.
Given a memory table trained with a source model~$A$, we freeze the table, attach it to a different target model~$B$, and train only a lightweight target-side reader for memory extraction.
If the frozen memory still improves the target after the original host is removed, then it contains usable information beyond source-specific co-adaptation.
If it fails, then the apparent memory benefit was likely tied to the original backbone.

\Cref{fig:transfer-overview} illustrates the life cycle of memory reuse and transfer: source-side memory training, memory export and freezing, and target-side extraction through a new reader.
To make this test precise, we separate three roles: (1) \emph{addressing} decides which entries are read, (2) \emph{memory} denotes the stored vectors, and (3) \emph{reader} maps them into the model.
We standardize addressing, freeze the memory, and adapt only the reader.
This turns portability into a measurable question: can a target model learn to read a fixed memory artifact?
We use a tokenizer-agnostic canonicalization pipeline to keep memory addresses stable across tokenizers, freeze the exported source memory, and train only a small target-side reader that projects memory values into the target residual stream.

Our central claim is: \textbf{for portable hashed external memory, memory presence alone is not enough; successful reuse depends on the interface through which a target backbone addresses and consumes the stored representations.} Nevertheless, memory content still matters.
Transferred memory clearly outperforms permuted-key controls and reaches substantially better downstream Question Answering (QA) performance than random memory under the final evaluation protocol.
But the amount of usable signal depends on how well the target model can align and integrate what it retrieves.
This view explains why imperfect memory can still help with a capable reader, and why reader design can matter as much as retraining or enlarging the stored table in this regime.

We support this claim with a concise chain of evidence, spanning diverse model families and settings.
First, a source--target transfer study shows that frozen memory consistently improves target models across different model families and scales, including Pythia \citep{biderman2023pythia}, Qwen \cite{qwen2025qwen3}, TinyLlama \citep{zhang2024tinyllama}, Phi \cite{abouelenin2025phi}, LLaMA \citep{touvron2023llama}, and Mistral \citep{jiang2023mistral7b}, with relative perplexity reductions of up to 15.7\%.
A supplementary peer-to-peer setting further confirms that transfer remains beneficial in both directions between models of similar scale.
Second, ablation experiments show that transferred memory clearly outperforms the permuted-key and no-memory controls, while scratch-trained target memory remains competitive under the same target-side adaptation budget.
This suggests that memory reuse is a joint property of the stored content and the target side reader that makes it usable.
Third, on downstream tasks, upgrading the target-side reader to a dual-layer, multi-branch architecture nearly closes the gap between same-model and cross-model reuse, with the best configuration reaching an average score of 38.8 across five QA tasks under our controlled evaluation protocol, while broader knowledge-intensive benchmarks show selective gains.

In summary, our contributions are:
\begin{itemize}
    \item We formulate cross-model frozen-memory reuse as an evaluation problem for external memory, separating portable memory from backbone-specific co-adaptation and measuring whether the target model can extract usable signal from a frozen artifact.
    \item We instantiate this problem with a transfer protocol for Engram-style memory that standardizes addressing across tokenizers, and we propose lightweight target-side readers for memory extraction, including stronger multi-branch, multi-layer variants.
    \item We provide an evidence chain showing that reader design is first-order: consistent gains across diverse model families and settings, ablations that isolate memory content and address integrity, and controlled QA results showing that stronger target-side readers, rather than source-backbone identity alone, drive the best performance.
\end{itemize}

This paper is organized as follows.
\Cref{sec:background} reviews Engram architecture, related memory mechanisms, and cross-model alignment.
\Cref{sec:method} describes the cross-model frozen-memory transfer protocol, including tokenizer-agnostic addressing, target-side reader design, and training regime.
\Cref{sec:experiments} presents the experimental setup, research questions, results, and takeaways.
Supplementary material in \Cref{app:sec:method_details} provides additional details on canonicalization, memory export, and the reader implementation.
\Cref{app:sec:implementation_details} gives training and evaluation details, including hyperparameters, datasets, and evaluation metrics.
\Cref{app:sec:transferability} extends RQ1 in~\Cref{sec:exp:transferability} with exact matrix values and controls for tokenizer mismatch, peer-scale transfer, target scaling, and representational similarity.
\Cref{app:sec:downstream_boundaries} extends RQ3 in~\Cref{sec:exp:downstream_boundaries} by diagnosing why downstream gains are task-dependent, separating source-corpus specialization, Phase-2 reader-fitting alignment, mixed-corpus behavior, and out-of-domain effects.
The supplementary material concludes with a computational cost analysis (\Cref{app:sec:computational_cost}), broader ethics and impact considerations (\Cref{app:sec:ethics}), and limitations (\Cref{app:sec:limitations}).

\section{Background and Related Work}
\label{sec:background}

\subsection{Engram Architecture}
\label{sec:bg:engram}

Engram~\citep{cheng2026engram} augments a Transformer-MoE backbone with an external conditional memory.
It retrieves static $n$-gram embeddings through deterministic hashing and injects the retrieved vector into the backbone through a learned gate.
The key property for our setting is that Engram separates \emph{addressing}, \emph{storage}, and \emph{read}: the input text determines which memory entry is read, the memory table stores the external vectors, and a small learned reader maps the retrieved vector into the backbone hidden space.
This separation makes transfer possible: as long as another model can reproduce the same memory addresses through a shared addressing scheme, the frozen Engram table can be reused through a target-side reader.

For each $n$-gram order $n \in [2,N]$ and hash head $k \in [1,K]$, a deterministic hash $\varphi_{n,k}$ maps the canonicalized $n$-gram $g_{t,n}$ to an entry in memory table $\mathbf{E}_{n,k}$.
The retrieved embeddings are concatenated as
\begin{equation}
    \mathbf{e}_t \triangleq 
    \bigoplus_{n=2}^{N} 
    \bigoplus_{k=1}^{K} 
    \mathbf{E}_{n,k}[\varphi_{n,k}(g_{t,n})].
    \label{eq:engram-retrieval}
\end{equation}
Crucially, this lookup depends on the canonicalized input sequence rather than the model hidden state, so the memory table can be treated as an external artifact once the address space is standardized.

The retrieved vector is consumed through learned projections and a gate:
\begin{equation}
\begin{aligned}
    s_t &=
    \frac{\RMSNorm(\mathbf{h}_t)^\top \RMSNorm(\mathbf{k}_t)}{\sqrt{d}},
    &
    \alpha_t &= \sigma\!\left(\mathrm{sign}(s_t)\sqrt{|s_t|}\right), \\
    \mathbf{k}_t &= \mathbf{W}_K \mathbf{e}_t,
    &
    \mathbf{v}_t &= \mathbf{W}_V \mathbf{e}_t .
\end{aligned}
    \label{eq:engram-gate}
\end{equation}
Here, $\mathbf{W}_K$, $\mathbf{W}_V$, and the gate define the backbone-specific reader.
This structural separation makes frozen-memory transfer possible: the source memory table can be frozen and moved to a target model, while only a lightweight target-side reader is trained to extract and align it.
\subsection{Memory Mechanisms}
\label{sec:bg:memory}

MoE~\citep{shazeer2017outrageously, fedus2022switch, deepseek2024moe} expands capacity through conditional computation, but its useful signal remains embedded in the host backbone rather than exposed as a reusable memory artifact.
Other memory-augmented models also miss the extraction question for one of two reasons: they either retrieve from large dynamic memory banks at inference time~\citep{graves2014ntm, weston2015memory}, or they store information in model-specific hidden spaces, as in KNN-LM~\citep{khandelwal2021generalization}.
Retrieval-augmented generation and RETRO-style systems~\citep{lewis2020rag, borgeaud2022improving, izacard2023atlas} are portable at the document level, but the reusable object is a text datastore rather than a trained memory representation.
Larger learned memory layers~\citep{lample2019large, berges2025memory, behrouz2025titans, chen2025mlpmemory} improve capacity or efficient knowledge access, but their stored representations are usually trained and consumed within the same model.

Engram differs in the regime relevant here: deterministic hashing provides an explicit address space, the memory table is external and addressable, and a small learned reader mediates read by the backbone.
This separation makes a frozen memory artifact well defined and allows portability to be tested as extraction rather than as additional model capacity.
A broader qualitative comparison appears in \Cref{tab:comparison}.

\subsection{Cross-Model Alignment}
\label{sec:bg:alignment}

Cross-model transfer is often studied as representation or adapter alignment.
Embedding-space work suggests that independently trained models can sometimes be connected by simple linear maps between hidden representations~\citep{mikolov2013exploiting, kornblith2019similarity, huh2024platonic, chen2025stitching}.
PEFT transfer methods such as Cross-LoRA~\citep{xia2025crosslora}, LoRA-X~\citep{zhang2025lorax}, and Trans-LoRA~\citep{wang2025translora} show that compact adapters or low-rank updates can sometimes be reused across heterogeneous models.
These methods are useful baselines for parameter transfer, but they do not directly address frozen external-memory reuse: there is no separate memory table to address, freeze, and extract from after the source backbone is removed.
In frozen-memory extraction, alignment starts earlier.
The target model must first read the same memory entry as the source-intended lookup, even when their tokenizers differ.
Otherwise, the same surface text may be routed to unrelated memory slots, and any downstream hidden-space mapping becomes ill-defined.
Our method, therefore, treats alignment as a two-stage problem: tokenizer-agnostic canonicalization aligns the key space, and a lightweight target-side reader aligns the retrieved value space to the target residual stream.

\section{Methods}
\label{sec:method}
\begin{figure}[ht]
    \centering
    \includegraphics[width=\columnwidth]{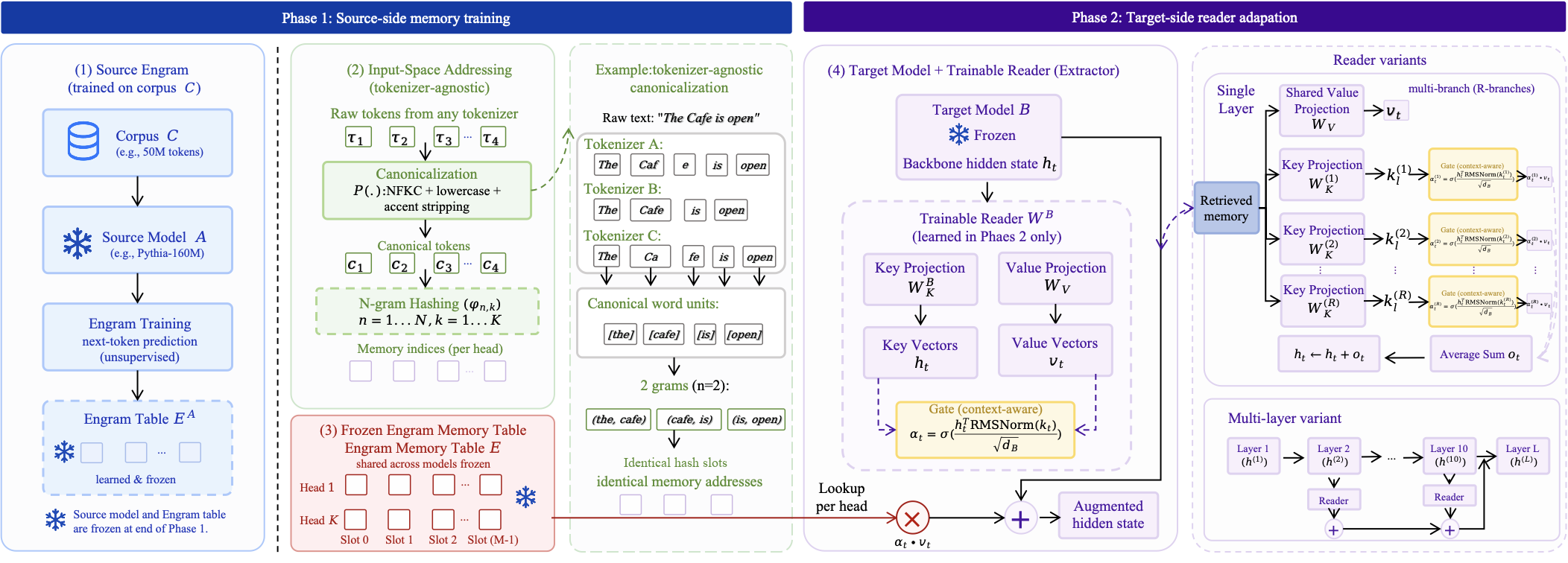}
    
    \caption{Overview of the cross-model memory transfer.
    A memory table trained with source model~$A$ is frozen and attached to target model~$B$.
    Canonicalization keeps the address space fixed across tokenizers, and only the target-side reader (including multi-head and multi-layer variants) is trained.
    This setup operationalizes whether external memory remains reusable outside its source backbone.}
    \label{fig:transfer-overview}
\end{figure}

Our method recasts Engram as a transferable artifact.
We freeze the source memory and fit only a small target-side reader.
This requires a shared address space, frozen memory, and a lightweight reader.
Relative to native Engram, we make three changes: we (1) freeze the exported source memory, (2) replace model-specific token-ID lookup with tokenizer-agnostic canonicalization, and (3) use a lightweight target-side reader attached by residual injection to read the knowledge from Engram memory.
\Cref{fig:transfer-overview} illustrates the overall architecture and training and adaptation procedures, and \Cref{alg:transfer} summarizes the training procedure.

\subsection{Tokenizer-Agnostic Addressing}
\label{sec:method:canon}

The first obstacle to transfer is the address mismatch.
Native Engram hashes model-specific token IDs inside the host model, which is sufficient when storage and read happen under one tokenizer, but does not define a stable cross-model address space.
If source and target tokenizers segment the same text differently, then identical surface content may be routed to different memory entries.
To avoid this failure mode, we build the lookup space on canonicalized decoded text rather than raw token IDs.
Let $\mathcal{T}_B$ be the tokenizer of target model $B$, and let $s_{1:t} = \mathrm{decode}(\mathcal{T}_B(x_{1:t}))$ denote the decoded target prefix up to position $t$.
We normalize that prefix with a shared canonicalization function $\mathcal{P}$ and extract the completed word sequence
\begin{equation}
    (w_1,\ldots,w_{m(t)}) = \mathrm{words}\!\left(\mathcal{P}(s_{1:t})\right),
    \label{eq:canon-words}
\end{equation}
where $\mathrm{words}(\cdot)$ returns the canonicalized word-boundary units implied by the current decoded prefix.
Word-boundary addressing is the default for the English-like settings above, but it degenerates for scripts without reliable whitespace segmentation.
For such inputs, we replace word units with NFKC-normalized, case-folded non-whitespace Unicode character events and hash character spans using the same deterministic addressing interface.
\Cref{sec:multi-lingual} evaluates this variant on Chinese and Japanese.
In practice, $\mathcal{P}$ applies NFKC normalization, lowercasing, accent stripping, and whitespace cleanup.
Memory lookup is then performed over canonical word-boundary $n$-grams rather than model-specific token IDs:
\begin{equation}
    \mathbf{e}_t =
    \bigoplus_{n=2}^{N}
    \bigoplus_{j=1}^{K}
    \mathbf{E}_{n,j}\!\left[\varphi_{n,j}(w_{m(t)-n+1}, \ldots, w_{m(t)})\right].
    \label{eq:transfer-lookup}
\end{equation}
At subword positions inside the same decoded word, the current token reuses the same right-edge word context until a new word boundary appears.
The retrieved vector $\mathbf{e}_t$ is therefore determined by canonicalized text and the frozen memory artifact, not by target-specific token IDs.
Two target models with different tokenizers can still read the same transferred entry as long as they share the same canonicalization and hash specification.

\begin{algorithm}[ht]
\caption{Cross-Model Frozen-Memory Extraction}
\label{alg:transfer}
\small \KwIn{Source memory $\mathcal{E}_A$, target model $B$, target corpus $\mathcal{D}$, selected injection layers $\mathcal{L}$, branch count $R$, canonicalization $\mathcal{P}$} \KwOut{Target model $B$ augmented with frozen transferred memory} \BlankLine \tcp{\textbf{// Phase 1 artifact is already trained; prepare transfer}} \textbf{Extract} Engram tables $\{\mathbf{E}_{n,j}\}$ and hash specification from $\mathcal{E}_A$\; \textcolor{algred}{\textbf{Freeze} all transferred tables $\{\mathbf{E}_{n,j}\}$ \hfill \textit{[portable memory artifact]}}\; \BlankLine \tcp{\textbf{// Initialize target-side reader and hooks}} \ForEach{$\ell \in \mathcal{L}$}{ \textbf{Initialize} reader parameters $\mathcal{W}^{(B,\ell)}$\; \textbf{Set} $\mathcal{W}^{(B,\ell)} \leftarrow \{\mathbf{W}_{K,\ell,r}^{(B)}, \beta_{\ell,r}\}_{r=1}^{R}, \mathbf{W}_{V,\ell}^{(B)}$\; \If{$R=1$}{ \textbf{Set} $\beta_{\ell,1} \leftarrow 0$\;
    }
    \textbf{Register} a forward hook at layer $\ell$ that computes $\mathbf{o}_{t,\ell}$ from the current $\mathbf{h}_{t,\ell}$ and looked-up $\mathbf{e}_t$, then injects $\mathbf{h}_{t,\ell} \leftarrow \mathbf{h}_{t,\ell} + \mathbf{o}_{t,\ell}$\;
}
\BlankLine \tcp{\textcolor{algblue}{\textbf{// Phase 2: Reader-only adaptation}}} \For{$x_{1:T} \in \mathcal{D}$}{ \textbf{Canonicalize} each decoded prefix into shared word-boundary $n$-grams with $\mathcal{P}$\; \textbf{Look up} frozen memory vectors $\mathbf{e}_{1:T}$ from $\{\mathbf{E}_{n,j}\}$\; \textbf{Run} the frozen backbone on $x_{1:T}$ with the registered hooks active\; \textcolor{algblue}{\textbf{Update only} reader parameters $\{\mathcal{W}^{(B,\ell)}\}_{\ell \in \mathcal{L}}$ using the LM loss \hfill \textit{[backbone and memory frozen]}}\;
}
\end{algorithm}

\subsection{Target-Side Reader}
\label{sec:method:adaptor}

Once the address space is shared, the remaining problem is extraction: how should the target model consume a memory vector whose geometry was shaped by a different backbone?
We use a shared-value reader family parameterized by branch count $R$, where a larger $R$ increases target-side alignment capacity without increasing memory size.
For branch $r \in \{1, \ldots, R\}$ at layer $\ell$:
\begin{equation}
    \mathbf{k}_{t,\ell}^{(r)} = \mathbf{W}_{K,\ell,r}^{(B)} \mathbf{e}_t,
    \qquad
    \mathbf{v}_{t,\ell} = \mathbf{W}_{V,\ell}^{(B)} \mathbf{e}_t,
    \label{eq:adaptor-branch-proj}
\end{equation}
\begin{equation}
    \alpha_{t,\ell}^{(r)} = \sigma\!\left(
    \frac{\RMSNorm(\mathbf{h}_{t,\ell})^\top \RMSNorm(\mathbf{k}_{t,\ell}^{(r)})}{\sqrt{d_B}}
    + \beta_{\ell,r}
    \right),
    \label{eq:adaptor-branch-gate}
\end{equation}
\begin{equation}
    \mathbf{o}_{t,\ell} =
    \frac{1}{R}\sum_{r=1}^{R}
    \alpha_{t,\ell}^{(r)} \cdot \mathbf{v}_{t,\ell}.
    \label{eq:adaptor-branch-agg}
\end{equation}
The resulting reader output is injected through a residual connection:
\begin{equation}
    \mathbf{h}_{t,\ell} \leftarrow \mathbf{h}_{t,\ell} + \mathbf{o}_{t,\ell}.
    \label{eq:adaptor-inject}
\end{equation}
The shared value projection forces all branches to operate on the same retrieved content and differ only in \emph{how} they gate and align it.
The added capacity is therefore an extraction mechanism rather than extra storage.
The main $3 \times 3$ transfer matrix uses $R=1$; the stronger QA reader uses $R=4$.

\smallskip\noindent\textbf{Multi-layer injection.} The same reader family can be instantiated at one or more target layers $\mathcal{L}$.
Single-layer injection is sufficient for the cross-architecture transfer matrix, whereas the best QA configuration uses dual-layer injection at layers 2 and 10 with $R=4$ branches.

\subsection{Training Regime}
\label{sec:method:training}

The training objective is the standard next-token language modeling loss on the target corpus.
During target-side adaptation, both the exported source memory tables $\{\mathbf{E}_{n,j}\}$ and the target backbone parameters $\theta_B$ remain frozen; only the reader parameters $\{\mathcal{W}^{(B,\ell)}\}_{\ell \in \mathcal{L}}$ are updated.
Storage is fixed, and only extraction is learned.
Algorithm~\ref{alg:transfer} summarizes the procedure concisely.
The key operations are to freeze the exported source memory, canonicalize target inputs into the shared address space, and update only the target-side reader.
\paragraph{Deployment modes.}
The two-phase protocol above describes the general case in which the target requires a model-specific reader.
We additionally distinguish between two deployment modes.
In \emph{direct artifact reuse}, a provider exports the frozen memory together with a reader whose residual interface is compatible with the consumer model; the consumer performs no target-side optimization.
In \emph{reader-adapted reuse}, the same frozen memory is retained, but a lightweight target-specific reader is fitted on an available adaptation stream.
The first mode minimizes consumer-side cost, while the second provides additional alignment capacity when direct reader reuse is insufficient.
Both modes preserve the defining property of the protocol: the transferred memory table itself is never rewritten.

\section{Experiments and Results}
\label{sec:experiments}


In this section, we address the effect of cross-model memory transfer through six research questions.
All experiments follow the same protocol: a source memory is trained (Phase 1), then frozen and attached to a target model where only a reader is trained (Phase 2) as illustrated in \Cref{fig:transfer-overview}.
We evaluate on intrinsic language modeling and downstream QA tasks.
Intrinsic evaluations use the memory transfer architecture with a minimal reader, while QA evaluation uses a stronger multi-layer, multi-branch reader.
Implementation details are in \Cref{app:impl}.

\subsection{RQ1: Does frozen memory transfer across backbones, tokenizers, and scales?}
\label{sec:exp:transferability}

Our first question is whether the exported memory remains useful once detached from the source backbone.
We start with a $3 \times 3$ source--target matrix spanning three source memories (Pythia-160M ~\citep{biderman2023pythia}, Qwen3.5-0.8B, Qwen3.5-9B~\citep{qwen2025qwen3}) and three target models (Pythia-410M, Qwen3.5-4B, TinyLlama-1.1B~\citep{zhang2024tinyllama}), all trained with the same 20M-token target-side protocol.
\Cref{app:sec:transferability} therefore stress-tests the same conclusion against several alternative explanations.
It reports the exact matrix values, introduces parameter-matched and retrieval-based comparisons, removes the small-to-large transfer asymmetry through peer-to-peer transfer, tests whether gains persist as the target scales, and examines whether representational similarity predicts transfer success.
\begin{figure*}[ht]
\centering
\begin{subfigure}[t]{0.50\textwidth}
    \centering
    \includegraphics[width=\linewidth]{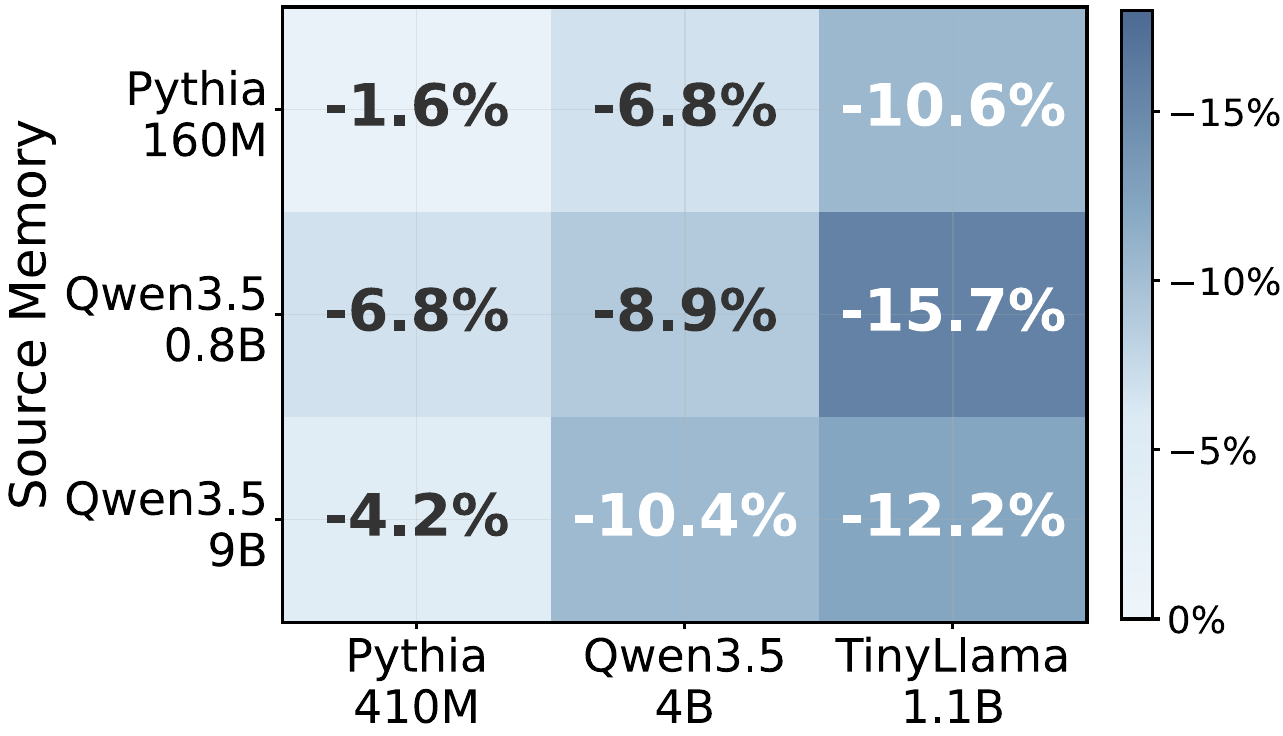}
    \caption{Full $3 \times 3$ source--target transfer matrix, reported as relative PPL reduction over the no-memory baseline.}
    \label{fig:transfer_matrix_heatmap}
\end{subfigure}
\hfill
\begin{subfigure}[t]{0.48\textwidth}
    \centering
    \includegraphics[width=\linewidth]{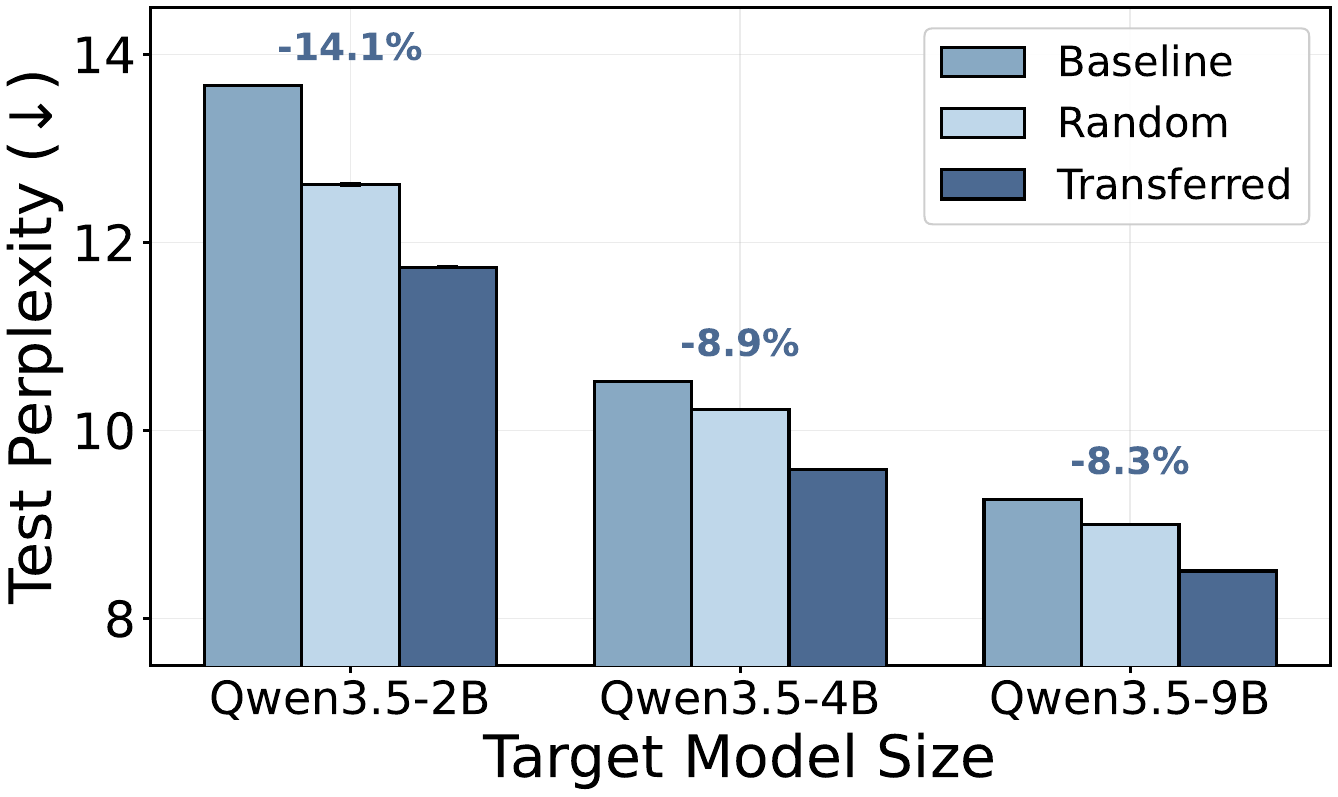}
    \caption{Target-scale intrinsic transfer: the same transferred Qwen3.5-0.8B memory improves Qwen3.5 targets from 2B to 9B.}
    \label{fig:target_scale_transfer}
\end{subfigure}
\caption{The full 3x3 transfer matrix and target-scale transfer performance across different model types trained on Wikitext-103.}
\label{fig:transfer_scaling_visual}
\end{figure*}

\Cref{fig:transfer_scaling_visual} summarizes this result together with the scaling trend.
All nine cells in the main matrix in \Cref{fig:transfer_matrix_heatmap} show PPL reduction in models with transferred memory over the no-memory baseline, and the same transferred Qwen3.5-0.8B memory continues to help as the target grows from 2B to 9B (\Cref{fig:target_scale_transfer}), compared with the baseline without memory and the random memory control with a randomly initialized, untrained memory of the same architecture and size as the transferred memory.

The full matrix in \Cref{fig:transfer_matrix_heatmap} gives a positive answer in every one of the nine cells.
Transferred memory improves over the no-memory baseline in all source--target pairs, with relative gains from 1.6\% to 15.7\%.
The strongest gain appears in the Qwen3.5-0.8B $\to$ TinyLlama-1.1B case, showing that large improvements do not require the source and target to share an architecture or a tokenizer.
The same cell also gives a 13.9\% reduction with random memory, so the 15.7\% headline should not be interpreted as a 15.7\% content-specific gain; the transferred table adds a smaller but positive margin over the random control.

This is also not only an intrinsic-perplexity effect: \Cref{fig:downstream_tasks} later shows selective downstream gains for Qwen3.5 targets from 0.8B to 9B, and \Cref{tab:domain} shows that the same target family retains that pattern under two different source-corpus families.
On Qwen3.5-2B specifically, \Cref{fig:qwen_efficiency_visual} shows the same asymmetry in a more focused form: transferred memory stays ahead of scratch on intrinsic PPL across all three budgets (\Cref{fig:qwen2b_ppl_scaling}), while the downstream gains are selective rather than universal, with the clearest best-of-budget improvements appearing on BoolQ, RTE, and SciQ (\Cref{fig:qwen2b_qa_scaling}).

One subtle pattern is that source scaling helps, but not monotonically, for every target.
Qwen3.5-0.8B improves over Pythia-160M on all three targets, yet Qwen3.5-9B does not always beat Qwen3.5-0.8B on the most dissimilar target (Pythia-410M), suggesting that transfer depends not only on what the source memory stores, but also on what the target reader can successfully extract.
This underscores the importance of the reader's extraction capability, as supplementary results in \Cref{app:sec:transferability} confirm that transfer is effective in peer-to-peer settings (\Cref{sec:exp:peer}) and even between backbones with low representational similarity (\Cref{sec:exp:cka}).

\takeawaybox{Frozen memory transfer remains effective across backbone families, tokenizer boundaries, and target scales, with gains in all nine main source--target pairs; the same Qwen3.5 target family also retains selective downstream gains beyond perplexity, and successful transfer ultimately depends on whether the target reader can extract the stored structure.}

\subsection{RQ2: How much does reader design matter?}
\label{sec:exp:reader_design}

The previous experiments established that memory transfer remains effective even across model families.
However, they do not explain where the transfer gains actually come from.
One possibility is that most of the benefit is already contained in the transferred memory itself.
An alternative explanation is that transfer quality is primarily limited by the target-side reader, i.e., the mechanism responsible for extracting and integrating information from the frozen memory.

To distinguish between these possibilities, we keep the transferred memory fixed and vary only the reader architecture.
Specifically, we study two aspects of reader design: (i) memory placement, namely the layers at which memory is injected into the target backbone, and (ii) retrieval capacity, controlled through the number of reader branches.
Our central finding is that reader design is a first-order factor: most of the gain comes from placing the reader at the right layers, while multiple branches further improve performance by accessing the same memory from different perspectives.

\begin{table*}[!htbp]
\centering
\caption{
QA accuracy (\%) on Mistral-7B-v0.3 for different reader variants and baselines.
Unless noted otherwise, Engram and baseline rows report F1; the MLP Memory row reproduces the source paper's reported Exact Match and is included only as a non-comparable reference.
Budget notation follows \Cref{tab:engram_QA_summary_new} trained on Wikipedia-2021~\citep{izacard2023atlas}.
}
\label{tab:QA_mistral_summary}
\footnotesize
\renewcommand{\arraystretch}{1.15}
\setlength{\tabcolsep}{2.4pt}
\begin{tabular}{lccccccc}
\toprule
\textbf{Variant} & \textbf{Budget} & \textbf{NQ} & \textbf{WebQA} & \textbf{TriviaQA} & \textbf{TruthQA} & \textbf{HotpotQA} & \textbf{Average} \\
\midrule
Base & -- & 20.6 & 29.3 & 57.7 & 32.1 & 21.0 & 32.1 \\
\rowcolor{groupgray}
\multicolumn{8}{c}{\textit{Non-parametric Methods}} \\
RAG~\citep{lewis2020rag} & -- & 22.6{\scriptsize\,\pos{1.9}} & 24.9{\scriptsize\,\negc{-4.4}} & 54.2{\scriptsize\,\negc{-3.4}} & \textbf{35.5}{\scriptsize\,\pos{3.4}} & \textbf{29.8}{\scriptsize\,\pos{8.8}} & 33.4{\scriptsize\textcolor{posgreen}{(+3.9\%)}} \\
kNN~\citep{khandelwal2021generalization} & -- & 21.1{\scriptsize\,\pos{0.4}} & 30.5{\scriptsize\,\pos{1.2}} & 57.8{\scriptsize\,\pos{0.1}} & 32.3{\scriptsize\,\pos{0.2}} & 21.2{\scriptsize\,\pos{0.2}} & 32.6{\scriptsize\textcolor{posgreen}{(+1.4\%)}} \\
\rowcolor{groupgray}
\multicolumn{8}{c}{\textit{Parametric Methods}} \\
CPT & -- & 12.2{\scriptsize\,\negc{-8.5}} & 34.1{\scriptsize\,\pos{4.8}} & 61.2{\scriptsize\,\pos{3.6}} & 29.2{\scriptsize\,\negc{-2.9}} & 16.0{\scriptsize\,\negc{-4.9}} & 30.5{\scriptsize\textcolor{negred}{(-5.0\%)}} \\
LoRA~\citep{hu2022lora} & -- & 18.2{\scriptsize\,\negc{-2.5}} & 34.5{\scriptsize\,\pos{5.2}} & 61.6{\scriptsize\,\pos{4.0}} & 30.9{\scriptsize\,\negc{-1.2}} & 16.2{\scriptsize\,\negc{-4.7}} & 32.3{\scriptsize\textcolor{posgreen}{(+0.5\%)}} \\
MLP Memory~\citep{chen2025mlpmemory} & -- & 25.2{\scriptsize\,\pos{4.6}} & \textbf{37.5}{\scriptsize\,\pos{8.2}} & 61.0{\scriptsize\,\pos{3.3}} & 32.5{\scriptsize\,\pos{0.5}} & 24.1{\scriptsize\,\pos{3.2}} & 36.1{\scriptsize\textcolor{posgreen}{(+12.3\%)}} \\
\rowcolor{groupgray}
\multicolumn{8}{c}{\textit{Engram Variants (sorted by Average)}} \\
LLaMA Frozen $\{10\}$-R1 & 10/0 & 21.8{\scriptsize\,\pos{1.2}} & 29.6{\scriptsize\,\pos{0.3}} & 60.6{\scriptsize\,\pos{2.9}} & 33.1{\scriptsize\,\pos{1.1}} & 22.3{\scriptsize\,\pos{1.4}} & 33.5{\scriptsize\textcolor{posgreen}{(+4.4\%)}} \\
Mistral Frozen $\{10\}$-R1 & 10/0 & 21.9{\scriptsize\,\pos{1.2}} & 29.7{\scriptsize\,\pos{0.4}} & 60.8{\scriptsize\,\pos{3.1}} & 32.9{\scriptsize\,\pos{0.8}} & 22.4{\scriptsize\,\pos{1.4}} & 33.5{\scriptsize\textcolor{posgreen}{(+4.4\%)}} \\
Mistral $\{10\}$-R1 & 10/20 & 22.1{\scriptsize\,\pos{1.5}} & 31.6{\scriptsize\,\pos{2.4}} & 61.1{\scriptsize\,\pos{3.4}} & 32.8{\scriptsize\,\pos{0.7}} & 22.3{\scriptsize\,\pos{1.3}} & 34.0{\scriptsize\textcolor{posgreen}{(+5.8\%)}} \\
LLaMA $\{10\}$-R1 & 10/20 & 22.2{\scriptsize\,\pos{1.5}} & 32.6{\scriptsize\,\pos{3.3}} & 61.2{\scriptsize\,\pos{3.6}} & 32.9{\scriptsize\,\pos{0.8}} & 22.3{\scriptsize\,\pos{1.3}} & 34.2{\scriptsize\textcolor{posgreen}{(+6.6\%)}} \\
LLaMA $\{2,10\}$-R1 & 10/20 & 30.3{\scriptsize\,\pos{9.7}} & 28.7{\scriptsize\,\negc{-0.6}} & \textbf{70.1}{\scriptsize\,\pos{12.4}} & 31.0{\scriptsize\,\negc{-1.1}} & 27.4{\scriptsize\,\pos{6.4}} & 37.5{\scriptsize\textcolor{posgreen}{(+16.7\%)}} \\
\rowcolor{harmonyblue}
LLaMA $\{2,10\}$-R4 & 10/20 & \textbf{30.3}{\scriptsize\,\pos{9.7}} & 33.7{\scriptsize\,\pos{4.4}} & 69.9{\scriptsize\,\pos{12.3}} & 30.9{\scriptsize\,\negc{-1.2}} & 27.6{\scriptsize\,\pos{6.7}} & \textbf{38.5}{\scriptsize\textcolor{posgreen}{(+19.9\%)}} \\
\bottomrule
\end{tabular}
\end{table*}

The first pattern is that a weak or frozen reader improves only a small fraction of the performance.
With a single injection layer at $\{10\}$, the frozen variants achieve only 33.5--33.5 average accuracy, a modest improvement over the 32.1 base model.
Training the same single-layer reader raises performance to 34.0--34.2, indicating that target-side adaptation already contributes noticeably even before increasing reader capacity.
This suggests that the transferred memory alone is insufficient; the target model must learn how to access it effectively.

The largest gain comes from reader placement.
Moving from LLaMA $\{10\}$-R1 to LLaMA $\{2,10\}$-R1 increases average accuracy from 34.2 to 37.5.
The improvement is broad rather than dataset-specific: NQ rises from 22.20 to 30.3, TriviaQA from 61.2 to 70.1, and HotpotQA from 22.3 to 27.4.
These results suggest that dual-layer injection provides the target model with multiple opportunities to integrate retrieved memory throughout the forward pass, instead of forcing all memory interaction through a single late-layer interface.

Reader capacity explains the remaining improvement.
Keeping the injection layers fixed at $\{2,10\}$ and increasing the number of branches from $R=1$ to $R=4$ raises average accuracy from 37.5 to 38.5.
The most visible gain appears on WebQA (28.7 to 33.7), while the strong improvements on NQ, TriviaQA, and HotpotQA are preserved.
This behavior is consistent with the intended role of the multi-branch design: different branches provide complementary key-gating pathways over the same shared value representation.
The branches, therefore, increase the flexibility with which the target conditions the contribution of a retrieved memory vector, rather than introducing independent stored value directions.

Reader structure is not the only capacity factor.
We additionally vary the frozen-memory interface width in a controlled Pythia-160M$\rightarrow$Pythia-410M diagnostic.
Increasing the memory width from 256 to 512 and 1024 reduces test perplexity from 22.368 to 22.157 and 21.808, respectively.
However, a width-512 reader injected at two layers reaches 21.522, outperforming the wider width-1024 single-layer configuration.
Thus, the default 512-dimensional interface is a design choice rather than a hard architectural bottleneck, and extraction capacity depends jointly on interface width and reader structure.
Because changing the width also changes both memory and reader parameter counts, we treat this experiment as a capacity-sensitivity analysis rather than a causal estimate of memory width alone.

The same table also places reader improvements in context relative to existing approaches.
The strongest transfer configuration, LLaMA $\{2,10\}$-R4, achieves 38.5 average accuracy, outperforming the included like-for-like baselines, including RAG, kNN-LM, CPT, and LoRA.
Notably, this improvement is obtained without modifying the transferred memory itself, reinforcing the conclusion that extraction quality, rather than memory storage, is the dominant factor governing transfer performance.

\takeawaybox{
Reader design is a first-order determinant of transfer quality.
Dual-layer injection recovers most of the available gain, while multi-branch gating provides an additional improvement by allowing the same retrieved value representation to be conditioned through multiple target-dependent key-gate pathways.
Together, these design choices raise average QA accuracy from 34.2 to 38.5 without changing the transferred memory itself.
}
\subsection{RQ3: When does transferred memory help downstream?}
\label{sec:exp:downstream_boundaries}

\begin{table*}[!htbp]
\centering
\caption{
QA accuracy (\%) on Mistral-7B-v0.3 for transfer verification and ablation studies.
Deltas are measured against the baseline.
Budget $=$ P1/P2 denotes source-memory and target-reader training tokens (millions)  trained on Wikipedia-2021.
}
\label{tab:engram_QA_summary_new}
\footnotesize
\renewcommand{\arraystretch}{1.15}
\setlength{\tabcolsep}{2.4pt}
\begin{tabular}{lccccccc}
\toprule
\textbf{Variant} & \textbf{Budget} & \textbf{NQ} & \textbf{WebQA} & \textbf{TriviaQA} & \textbf{TruthQA} & \textbf{HotpotQA} & \textbf{Average} \\
\midrule
Base & -- & 20.6 & 29.3 & 57.7 & 32.1 & 21.0 & 32.1 \\
\rowcolor{groupgray}
\multicolumn{8}{c}{\textit{Transfer Verification}} \\
\rowcolor{harmonyblue}
Mistral $\{2,10\}$-R4 & 10/20 & 30.2{\scriptsize\,\pos{9.5}} & \textbf{34.0}{\scriptsize\,\pos{4.7}} & 70.0{\scriptsize\,\pos{12.3}} & 30.8{\scriptsize\,\negc{-1.3}} & 27.7{\scriptsize\,\pos{6.8}} & \textbf{38.5}{\scriptsize\textcolor{posgreen}{(+20.0\%)}} \\
LLaMA $\{2,10\}$-R4 & 10/20 & 30.3{\scriptsize\,\pos{9.7}} & 33.7{\scriptsize\,\pos{4.4}} & 69.9{\scriptsize\,\pos{12.3}} & 30.9{\scriptsize\,\negc{-1.2}} & 27.6{\scriptsize\,\pos{6.7}} & 38.5{\scriptsize\textcolor{posgreen}{(+20.0\%)}} \\
Frozen Mistral $\{2, 10\}$-R4 & 10/0 & 30.5{\scriptsize\,\pos{9.8}} & 30.3{\scriptsize\,\pos{1.0}} & \textbf{70.8}{\scriptsize\,\pos{13.2}} & 32.2{\scriptsize\,\pos{0.2}} & 27.7{\scriptsize\,\pos{6.8}} & 38.3{\scriptsize\textcolor{posgreen}{(+19.2\%)}} \\
Frozen LLaMA $\{2, 10\}$-R4& 10/0 & \textbf{30.6}{\scriptsize\,\pos{10.0}} & 30.3{\scriptsize\,\pos{1.0}} & 70.8{\scriptsize\,\pos{13.1}} & 32.2{\scriptsize\,\pos{0.1}} & 27.6{\scriptsize\,\pos{6.7}} & 38.3{\scriptsize\textcolor{posgreen}{(+19.2\%)}} \\
\rowcolor{groupgray}
\multicolumn{8}{c}{\textit{Controls and Ablations}} \\
Mem-only Mistral & 10/0 & 0.2{\scriptsize\,\negc{-20.4}} & 0.3{\scriptsize\,\negc{-29.0}} & 1.4{\scriptsize\,\negc{-56.3}} & 33.0{\scriptsize\,\pos{0.9}} & 0.2{\scriptsize\,\negc{-20.7}} & 7.0{\scriptsize\textcolor{negred}{(-78.1\%)}} \\
Mem-only LLaMA & 10/0 & 0.0{\scriptsize\,\negc{-20.6}} & 0.0{\scriptsize\,\negc{-29.3}} & 0.0{\scriptsize\,\negc{-57.7}} & -- & 0.0{\scriptsize\,\negc{-21.0}} & -- \\
Permuted & 10/0 & 20.6{\scriptsize\,\pos{0.0}} & 28.9{\scriptsize\,\negc{-0.4}} & 63.1{\scriptsize\,\pos{5.4}} & 32.4{\scriptsize\,\pos{0.3}} & 23.1{\scriptsize\,\pos{2.2}} & 33.6{\scriptsize\textcolor{posgreen}{(+4.7\%)}} \\
FFN R4 & 10/20 & 19.6{\scriptsize\,\negc{-1.0}} & 27.3{\scriptsize\,\negc{-2.0}} & 58.1{\scriptsize\,\pos{0.5}} & 32.1{\scriptsize\,\negc{-0.0}} & 17.6{\scriptsize\,\negc{-3.4}} & 30.9{\scriptsize\textcolor{negred}{(-3.7\%)}} \\
\rowcolor{groupgray}
\multicolumn{8}{c}{\textit{Token Scaling (LLaMA2-7B source, $\{2,10\}$-R4)}} \\
Engram & 10/10 & 30.2{\scriptsize\,\pos{9.5}} & 29.9{\scriptsize\,\pos{0.6}} & 70.4{\scriptsize\,\pos{12.8}} & 31.2{\scriptsize\,\negc{-0.9}} & 27.6{\scriptsize\,\pos{6.7}} & 37.9{\scriptsize\textcolor{posgreen}{(+17.8\%)}} \\
Engram & 15/15 & 30.2{\scriptsize\,\pos{9.6}} & 32.7{\scriptsize\,\pos{3.4}} & 70.1{\scriptsize\,\pos{12.4}} & 30.6{\scriptsize\,\negc{-1.5}} & 27.8{\scriptsize\,\pos{6.8}} & 38.3{\scriptsize\textcolor{posgreen}{(+19.2\%)}} \\
Engram & 20/20 & 30.3{\scriptsize\,\pos{9.7}} & 33.9{\scriptsize\,\pos{4.6}} & 69.9{\scriptsize\,\pos{12.3}} & 30.7{\scriptsize\,\negc{-1.4}} & 27.6{\scriptsize\,\pos{6.7}} & 38.5{\scriptsize\textcolor{posgreen}{(+19.8\%)}} \\
Engram & 25/25 & 30.5{\scriptsize\,\pos{9.9}} & 33.1{\scriptsize\,\pos{3.8}} & 70.0{\scriptsize\,\pos{12.3}} & 30.9{\scriptsize\,\negc{-1.2}} & 27.7{\scriptsize\,\pos{6.7}} & 38.4{\scriptsize\textcolor{posgreen}{(+19.6\%)}} \\
\rowcolor{harmonyblue}
Engram & 30/30 & 30.6{\scriptsize\,\pos{10.0}} & 33.9{\scriptsize\,\pos{4.6}} & 70.4{\scriptsize\,\pos{12.7}} & 31.1{\scriptsize\,\negc{-1.0}} & \textbf{27.9}{\scriptsize\,\pos{7.0}} & \textbf{38.8}{\scriptsize\textcolor{posgreen}{(+20.7\%)}} \\
\bottomrule
\end{tabular}
\end{table*}
RQ1--RQ2 established that memory can be transferred across model families and that transfer quality is largely determined by the target-side reader.
The remaining question is not whether transfer works, but rather how far it can be pushed and where it begins to break down.

We study this question from two complementary perspectives.
First, we use QA benchmarks to estimate the upper bound of transfer performance under increasingly strong reader configurations.
Second, we evaluate whether these gains generalize beyond QA to broader knowledge-intensive tasks.
Throughout this section, $\{a,b\}$-R$r$ denotes memory injection at layers $\{a,b\}$ with $r$ retrieval branches, and ``Frozen'' indicates evaluation with the source reader without target-side adaptation.

\Cref{tab:engram_QA_summary_new} reveals a clear and surprisingly high transfer ceiling.
Under the strongest configuration ($\{2,10\}$-R4), same-model and cross-model transfer achieve nearly identical performance (38.5 vs.\ 38.5 average accuracy), achieving a ceiling once the reader is sufficiently expressive.
Performance further saturates with scale: increasing training budgets from 10/10 to 30/30 improves accuracy from 37.9 to 38.8, suggesting diminishing returns in the high-38 regime.

Control results confirm that this ceiling is not driven by parameter addition alone.
Memory-only variants collapse without a reader, and FFN-based substitutes fall below the base model, while permuting memory keys removes most gains.
These results indicate that successful transfer critically depends on the correct retrieval and integration of frozen memory rather than additional capacity.
The frozen-reader rows additionally expose a second deployment regime.
When the provider reader is directly compatible with the target residual interface, the complete frozen artifact can be reused without target-side training: mean QA remains approximately 38.3 compared with 32.1 for the no-memory target.
Optional target-reader fitting further raises the score to approximately 38.5.
Thus, target-side adaptation is useful but is not an inherent requirement for every compatible source--target pair.

We next examine whether these gains extend beyond QA to a broader set of downstream tasks.

\begin{figure}[ht]
    \centering
    \includegraphics[width=\columnwidth]{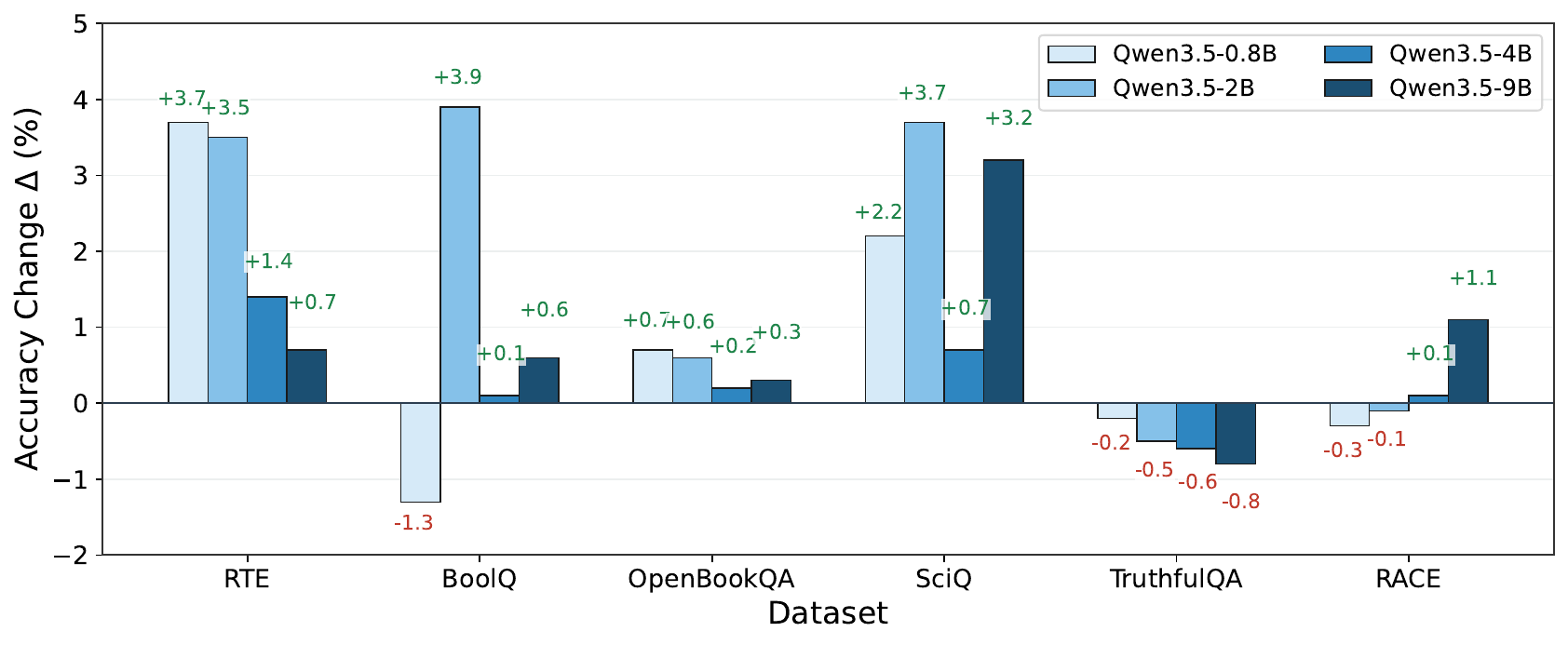}
    \caption{
    Downstream task evaluation.
    Bars report the change in accuracy ($\Delta$) produced by transferred memory relative to the no-memory baseline across six downstream tasks.
    Results are averaged over three seeds (42, 137, and 2024).
    Positive values indicate that transferred memory improves the target model after target-side adaptation on FineWeb-Edu (FW-9B)~\citep{lozhkovfineweb}.
    }
    \label{fig:downstream_tasks}
\end{figure}

\Cref{fig:downstream_tasks} shows that transfer benefits are selective rather than universal.
The high QA ceiling does not imply that transferred memory improves every downstream objective.
We therefore evaluate whether the benefits extend beyond the five QA benchmarks to a broader set of knowledge-intensive and reasoning tasks.
The strongest and most consistent gains appear on tasks for which additional factual or evidence-related information is likely to be useful.
RTE improves at every evaluated target scale, with gains ranging from $+0.7$ to $+3.7$ accuracy points.
SciQ also remains positive across scales, with improvements of up to $+3.7$ points.
OpenBookQA shows smaller but consistently positive gains, while BoolQ becomes positive for target models of 2B parameters and above.
Together, these results suggest that transferred memory is most useful when the downstream decision can benefit from additional factual associations or evidence support.

By contrast, the gains are substantially weaker on tasks whose success is less directly tied to factual retrieval. RACE remains close to zero across model sizes, suggesting that the imported memory contributes little to broad reading-comprehension performance when retrieving factual associations is not the primary bottleneck.

TruthfulQA provides the clearest negative boundary.
Performance decreases at every target scale, with changes ranging from $-0.2$ to $-0.8$ accuracy points.
Unlike the other evaluated tasks, TruthfulQA rewards calibrated truthfulness and resistance to commonly repeated misconceptions, rather than the recovery of frequently observed factual associations alone.
Additional memory-derived associations can therefore be unhelpful when the task requires the model to reject plausible but misleading information.
This result defines an important practical limit: transferred memory is most beneficial when the task rewards access to factual or evidence-related information, but it may be neutral or mildly harmful when success depends on calibration or resistance to memorized misconceptions.

Figure~\ref{fig:downstream_tasks} establishes this task-dependent pattern, but does not by itself identify why it arises.
The observed selectivity could reflect the information encoded in the source memory, the distribution used to fit the target-side reader during Phase~2, or a trade-off between specialization and broader distributional robustness.
To separate these factors, \Cref{app:sec:downstream_boundaries} presents a controlled diagnostic sequence.
It first varies source-corpus specialization while matching the Phase~2 fitting distribution, then holds the source memory fixed while varying the reader-fitting corpus.
It further examines whether mixed-corpus memory can broaden coverage without removing specialist gains, and finally evaluates whether the adapted reader introduces adverse effects on unrelated language-modeling distributions.

These supplementary experiments show that source-memory content and Phase~2 distribution alignment affect the magnitude of the realized gains, but do not remove the central downstream boundary.
Improvements remain concentrated on tasks that can exploit factual or evidence-related associations, and do not extend uniformly to broader reading comprehension, reasoning, or truthfulness-oriented evaluation.

\takeawaybox{
Memory transfer has a high practical ceiling: with a sufficiently expressive target-side reader, cross-model transfer nearly matches same-model reuse and saturates at approximately 38.5--38.8 average QA accuracy.
This ceiling, however, is task-dependent rather than universal. Transferred memory provides the clearest gains on factual and evidence-oriented tasks, has little effect on broader reading comprehension, and can be mildly harmful on TruthfulQA, where calibrated resistance to misleading associations is important.
}

\subsection{RQ4: Does transferred memory improve answer-token support in question answering?}
\label{sec:exp:answer_token_support}

Our fourth question is whether transferred memory measurably improves support for the correct answer, rather than merely changing final task accuracy in a hard-to-interpret way.
By \emph{answer-token support}, we mean the log-probability that the target model assigns to a valid gold-answer continuation at the positions where that answer is generated.
To test this, we compare four inference settings on the same LLaMA-2-7B $\to$ Mistral-7B dual-layer QA system used in \Cref{tab:qa_logit_analysis}: transferred memory, random memory, memory disabled, and an answer-position memory ablation.
We compute the length-normalized gold-answer log-probability for an alias $a$ as:
\begin{equation}
s_c(i,a) \triangleq \frac{1}{L_a} \sum_{t=1}^{L_a} \log p_c \left( y^{a}_{i,t} \mid x_i, y^{a}_{i,<t} \right),
\label{eq:score_alias}
\end{equation}
where $p_c$ is the target model distribution under condition $c \in \{\mathrm{trans}, \mathrm{rand}, \mathrm{disabled}, \mathrm{ablated}\}$, and $L_a$ is the token length of alias $a$.
The probability is computed with teacher forcing: the model is given the prompt and the gold-answer prefix, and we evaluate the log-probability assigned to the next gold-answer token.
For questions with multiple valid aliases $\mathcal{A}_i$, the question-level score is:
\begin{equation}
s_c(i) \triangleq \max_{a \in \mathcal{A}_i} s_c(i,a).
\label{eq:score_max}
\end{equation}
For datasets with a single annotated answer, $\mathcal{A}_i$ contains only that answer.
Finally, the mean score difference over the dataset $\mathcal{D}$ is:
\begin{equation}
\Delta \log p(c_1 - c_2) \triangleq \frac{1}{|\mathcal{D}|} \sum_{i \in \mathcal{D}} \left( s_{c_1}(i) - s_{c_2}(i) \right),
\label{eq:score_diff}
\end{equation}
where $\mathcal{D}$ is the evaluation set.
In \Cref{tab:qa_logit_analysis}, we report $\Delta \log p(\mathrm{trans}-\mathrm{rand})$, $\Delta \log p(\mathrm{trans}-\mathrm{disabled})$, and $\Delta \log p(\mathrm{trans}-\mathrm{ablated})$.
The first two differences ask whether transferred memory helps more than either an untrained table or no memory at all.
The third asks whether zeroing the memory channel only at answer-prediction positions reduces gold-answer likelihood.
For NQ and TriviaQA, where many examples have multiple valid aliases, the analysis scores all annotated aliases and keeps the best-valid gold continuation rather than forcing the first alias in the dataset.
\begin{table*}[!htbp]
\centering
\caption{QA memory-contribution trained on Wikipedia-2021 analysis for LLaMA-2-7B~\citep{touvron2023llama} $\to$ Mistral-7B-v0.3~\citep{jiang2023mistral7b} with dual-layer injection.
The $\Delta \log p$ columns report mean per-example differences in \emph{length-normalized} gold-answer log-probability.
Positive values indicate that transferred memory makes the gold answer more likely.}
\label{tab:qa_logit_analysis}
\footnotesize
\renewcommand{\arraystretch}{1.12}
\setlength{\tabcolsep}{2.7pt}
\begin{tabular}{lccccccc}
\toprule
\textbf{Task} & \textbf{Transferred} & \textbf{Random} & \textbf{Disabled} & \textbf{Ablated} & \textbf{\shortstack{$\Delta \log p$\\(Trans. - Random)}} & \textbf{\shortstack{$\Delta \log p$\\(Trans. - Disabled)}} & \textbf{\shortstack{$\Delta \log p$\\(Trans. - Ablated)}} \\
\midrule
NQ & 25.1 & 20.1 & 18.3 & 26.2 & \pos{+0.035} & \pos{+0.032} & \negc{-0.013} \\
WebQA & 32.3 & 27.4 & 31.1 & 33.6 & \pos{+0.147} & \pos{+0.197} & \negc{-0.001} \\
TriviaQA & 72.5 & 65.5 & 58.9 & 72.5 & \pos{+0.047} & \pos{+0.072} & \negc{-0.000} \\
TruthQA & 30.8 & 32.6 & 31.7 & 30.9 & \negc{-0.581} & \negc{-0.258} & \negc{-0.029} \\
HotpotQA & 27.1 & 22.9 & 18.6 & 27.2 & \pos{+0.022} & \pos{+0.019} & \negc{-0.009} \\
\bottomrule
\end{tabular}
\end{table*}

Transferred memory beats random memory and memory-disabled evaluation on four of the five QA subsets, both in task score and in gold-answer log-probability.
The cleanest gains appear on WebQA and TriviaQA.
On WebQA, transferred improves mean gold-answer log-probability by $+0.147$ over random memory and $+0.197$ over memory-disabled inference.
On TriviaQA, after fixing alias-aware gold-answer scoring, transferred also improves gold-answer log-probability by $+0.047$ over random memory and $+0.072$ over memory-disabled inference, aligning the logit-level analysis with the task-level F1 gains.
NQ and HotpotQA show the same direction with smaller positive margins.

TruthfulQA remains the main negative case.
Transferred is worse than both random memory and memory-disabled inference there, both in MC-average and in gold-answer log-probability.
This matches the broader boundary condition already visible in RQ3: transferred memory is most useful on factual retrieval-style QA, but not on truthfulness-style calibration.

The ablation column sharpens the interpretation.
For every task, transferred minus answer-position ablation is near zero or slightly negative rather than strongly positive.
That means the main benefit is not coming from a simple direct injection of the correct answer token at the final prediction positions.
Instead, the evidence points to an indirect effect: transferred memory improves the model's overall answer-supporting state on factual QA tasks, but direct answer-token memory contribution is weak and sometimes slightly harmful.

Taken together, RQ4 supports a narrower claim than ``memory use causes the gain''.
The gain is real on NQ, WebQA, TriviaQA, and HotpotQA, and the corrected alias-aware analysis strengthens that claim for TriviaQA in particular.
The results are therefore more consistent with improved factual support after target-side integration than with raw answer-position copying.

\takeawaybox{Transferred memory improves answer-token support selectively but positively: transferred memory increases gold-answer support on NQ, WebQA, TriviaQA, and HotpotQA relative to both random memory and no memory, with the corrected alias-aware TriviaQA analysis removing the main contradictory case, while the near-zero answer-position ablation effects indicate that the gain comes more from useful factual support than from direct answer-token injection.}


\subsection{RQ5: How data-efficient is frozen-memory transfer?}
\label{sec:exp:data_efficiency}

We first ask whether frozen-memory transfer reduces the amount of target-side
training required to reach competitive performance.
Across both Pythia and Qwen model pairs, transferred memory reaches strong
intrinsic performance substantially earlier than learning a new memory from
scratch, indicating improved target-side data efficiency.
The downstream gains are also positive, although their magnitude remains
task-dependent.

\paragraph{Controlled scaling on Pythia.}
\Cref{tab:scaling} provides a controlled comparison on
Pythia-160M $\rightarrow$ Pythia-410M by tracking the same transfer setting
across 5M, 20M, and 50M target-side training tokens.

\begin{table}[!htbp]
\scriptsize
\renewcommand{\arraystretch}{1.15}
\setlength{\tabcolsep}{2.5pt}
\centering
\caption{
Scaling analysis for frozen-memory transfer on
Pythia-160M $\rightarrow$ Pythia-410M.
Results report test perplexity as a function of target-side training tokens,
averaged over 3 seeds.
}
\label{tab:scaling}

\begin{tabular}{lcccc}
\toprule
\textbf{Condition} &
\textbf{Trainable Params} &
\textbf{5M} &
\textbf{20M} &
\textbf{50M} \\
\midrule

\rowcolor{harmonyblue}
\textbf{Transferred} &
\textbf{1.05M} &
$\mathbf{21.8}$ &
$\mathbf{21.5}$ &
$\mathbf{21.5}^{\dagger}$ \\

\rowcolor{groupgray}
From scratch &
34.6M &
$21.9$ &
$21.8$ &
$21.6$ \\

\bottomrule
\end{tabular}

\vspace{2pt}
\parbox{\linewidth}{
\scriptsize
$^{\dagger}$Early stopped at step 1{,}500 ($\approx$12M tokens);
validation PPL diverges beyond this point.
}
\end{table}

Transferred memory is already competitive after only 5M target-side tokens
and essentially saturates by 20M, reaching a test PPL of 21.5.
In contrast, training a new target memory from scratch improves more gradually
and approaches a comparable PPL only at the largest token budget.
This difference is particularly notable because the transferred condition
updates only 1.05M trainable parameters, compared with 34.6M for the
from-scratch condition, corresponding to roughly $3\%$ as many trainable
parameters.

The 50M transferred entry is reported at its best early-stopped checkpoint.
Although the nominal training budget is 50M tokens, validation perplexity
begins to diverge after approximately 12M tokens, and the best checkpoint
therefore matches rather than improves upon the 20M result.
Taken together, these results suggest that Phase~2 is primarily adapting and
reusing structure already learned in the source memory rather than relearning
an equivalent memory representation from the target corpus alone.

\begin{figure*}[ht]
\centering
\begin{subfigure}[t]{0.50\textwidth}
    \centering
    \includegraphics[width=\linewidth]{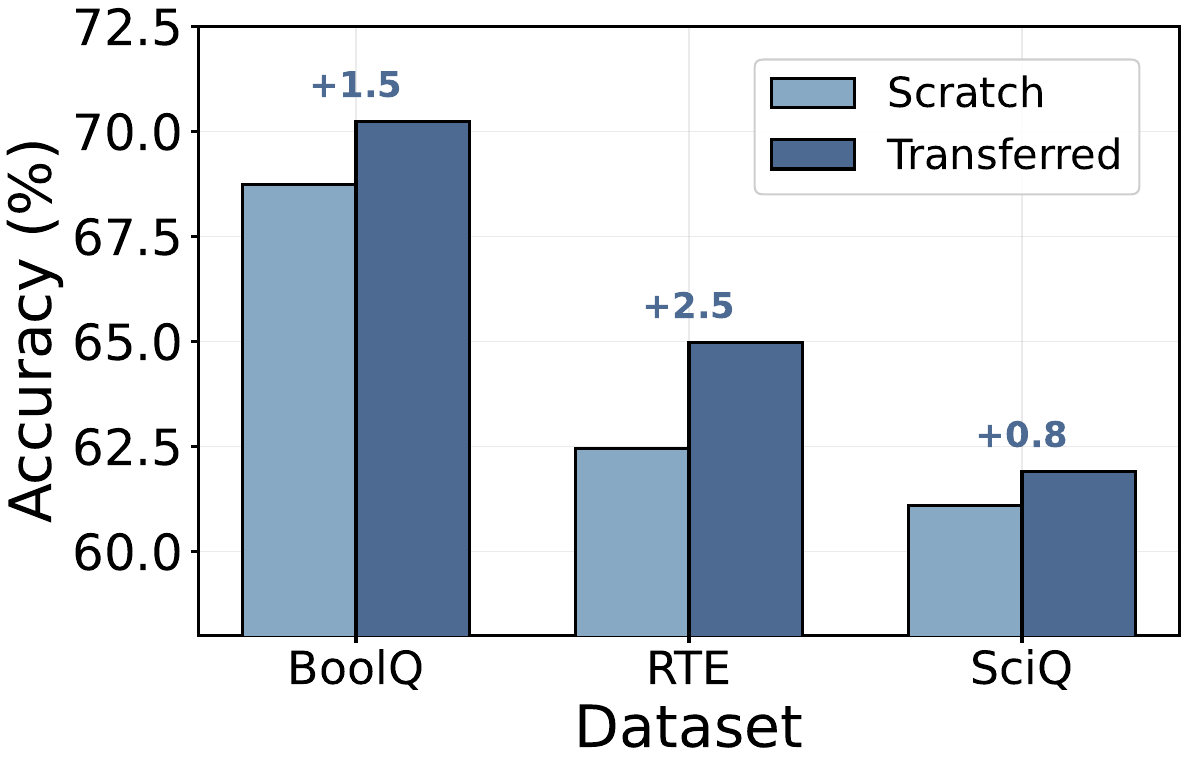}
    \caption{
    Qwen3.5-2B downstream QA scaling:
    average zero-shot accuracy across BoolQ, RTE, and SciQ as a function of
    target-side training tokens.
    }
    \label{fig:qwen2b_qa_scaling}
\end{subfigure}
\hfill
\begin{subfigure}[t]{0.49\textwidth}
    \centering
    \includegraphics[width=\linewidth]{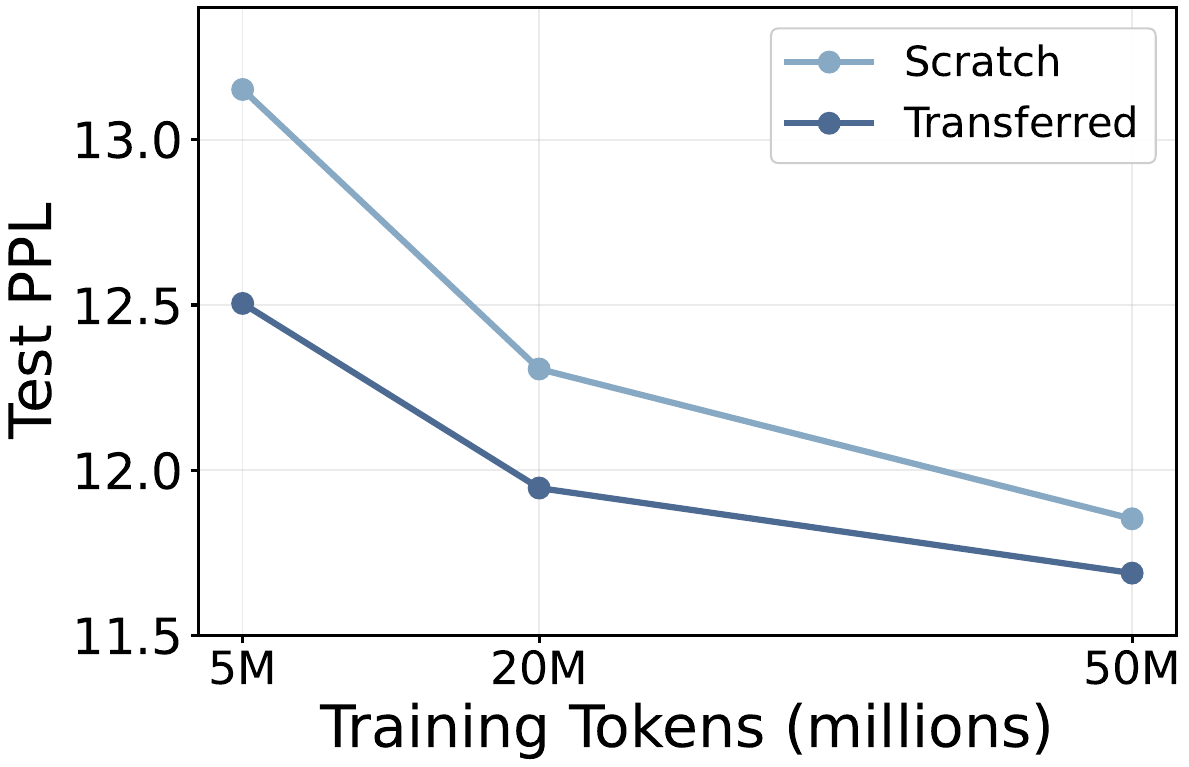}
    \caption{
    Qwen3.5-2B intrinsic scaling:
    test perplexity versus target-side training tokens for scratch and
    transferred memory.
    }
    \label{fig:qwen2b_ppl_scaling}
\end{subfigure}

\caption{
Target-side scaling for Qwen3.5-0.8B $\rightarrow$ Qwen3.5-2B with the
source memory trained on WikiText-103.
Transferred memory maintains a consistent intrinsic perplexity advantage,
while downstream improvements remain positive but task-dependent.
}
\label{fig:qwen_efficiency_visual}
\end{figure*}

\paragraph{Scaling to a larger target model.}
We next test whether the same trend persists at a larger scale using
Qwen3.5-0.8B $\rightarrow$ Qwen3.5-2B.
\Cref{fig:qwen_efficiency_visual} reports both intrinsic perplexity and
downstream QA performance as the amount of target-side training data
increases.

The Qwen results reinforce the Pythia scaling study.
Transferred adaptation achieves lower test perplexity than scratch training
across all evaluated target-side token budgets, showing that the intrinsic
data-efficiency advantage is not restricted to the smaller Pythia setting.
The downstream results show a similar but less uniform trend:
transferred memory improves BoolQ, RTE, and SciQ by $+1.5$, $+2.5$, and
$+0.8$ points, respectively.

These results also clarify the distinction between intrinsic adaptation and
downstream utility.
The perplexity advantage of transfer is stable across token budgets, whereas
the magnitude of downstream improvement depends on the task.
Thus, transferred memory provides a consistently better target-side
initialization, but the extent to which this advantage translates into
downstream accuracy still depends on whether the reader can extract and route
the stored signal in a task-compatible manner.

\takeawaybox{
Frozen-memory transfer is target-data efficient:
it reaches strong perplexity with substantially fewer target-side updates than
learning a new memory from scratch, while downstream gains remain positive but
task-dependent.
}


\subsection{RQ6: What drives the gains from memory transfer?}
\label{sec:exp:transfer_ablation}

Having established the target-side data-efficiency advantage of transfer,
we next ask what actually drives the downstream gains.
In particular, we test whether the improvement can be explained simply by
adding trainable target-side parameters, or whether successful transfer
depends on the transferred memory content, its addressing structure, and the
reader used to access it.

We study this question on
LLaMA-2-7B~\citep{touvron2023llama}
$\rightarrow$
Mistral-7B-v0.3~\citep{jiang2023mistral7b}
using the same dual-layer QA setting that produces our strongest downstream
results.
\Cref{tab:ablation} compares the full transferred reader against controls
designed to isolate the main components of the transfer mechanism.

``Transferred ($R{=}4$)'' denotes the full dual-layer four-branch reader.
``No gate'' removes context-dependent gating and always injects memory values.
``Affine stitch'' replaces the reader with a single affine
memory-to-hidden-state mapping.
``Permuted keys'' preserves the memory table but destroys addressing integrity
through index permutation.
``Random memory'' replaces the learned memory table with an untrained one.
``Train from scratch'' learns a fresh target memory under the same downstream
training setup.
Finally, ``FFN only'' replaces memory lookup with a parameter-matched residual
feed-forward module, while ``No memory baseline'' corresponds to the
underlying Mistral-7B model without memory augmentation.

\begin{table*}[!htbp]
\scriptsize
\renewcommand{\arraystretch}{1.15}
\setlength{\tabcolsep}{3.2pt}
\centering
\caption{
Ablation study of frozen-memory transfer on
LLaMA-2-7B $\rightarrow$ Mistral-7B-v0.3,
with the source memory trained on Wikipedia-2021.
Results report held-out test perplexity and downstream QA performance.
}
\label{tab:ablation}

\resizebox{\textwidth}{!}{
\setlength{\tabcolsep}{2pt}
\begin{tabular}{lcccccccc}
\toprule
\textbf{Condition} &
\textbf{Test PPL $\downarrow$} &
\textbf{95\% CI} &
\textbf{NQ} &
\textbf{WebQA} &
\textbf{TriviaQA} &
\textbf{TruthQA} &
\textbf{HotpotQA} &
\textbf{Avg $\uparrow$} \\
\midrule

\rowcolor{groupgray}
No memory baseline &
$9.9$ &
$[9.8, 10.1]$ &
$20.6$ &
$29.3$ &
$57.7$ &
$32.1$ &
$21.0$ &
$32.1$ \\

\rowcolor{harmonyblue}
\textbf{Transferred ($R{=}4$)} &
$8.7$ &
$[8.6, 8.9]$ &
$30.3$ &
$33.7$ &
$69.9$ &
$30.9$ &
$27.6$ &
$38.5$ \\

\rowcolor{groupgray}
\multicolumn{9}{c}{\textit{Interface Simplifications}} \\

No gate &
$10.1$ &
$[9.9, 10.3]$ &
$24.2$ &
$27.7$ &
$61.6$ &
$31.0$ &
$24.1$ &
$33.7$ \\

Affine stitch &
$9.5$ &
$[9.4, 9.7]$ &
$25.0$ &
$29.2$ &
$62.8$ &
$30.4$ &
$23.9$ &
$34.3$ \\

\rowcolor{groupgray}
\multicolumn{9}{c}{\textit{Content and Training Controls}} \\

Permuted keys &
$8.7$ &
$[8.5, 8.8]$ &
$18.3$ &
$34.5$ &
$57.7$ &
$30.6$ &
$21.4$ &
$32.5$ \\

Random memory &
$8.7$ &
$[8.6, 8.9]$ &
$23.0$ &
$27.1$ &
$65.1$ &
$33.6$ &
$24.1$ &
$34.6$ \\

Train from scratch &
$8.1$ &
$[8.0, 8.2]$ &
$29.9$ &
$33.0$ &
$70.4$ &
$30.9$ &
$27.9$ &
$38.4$ \\

\rowcolor{groupgray}
\multicolumn{9}{c}{\textit{No-memory Control}} \\

FFN only (param-matched $R{=}4$) &
$7.4$ &
$[7.2, 7.5]$ &
$25.5$ &
$27.9$ &
$64.1$ &
$31.6$ &
$23.6$ &
$34.5$ \\

\bottomrule
\end{tabular}%
}
\end{table*}

\paragraph{Parameter count alone does not explain the gains.}
The full transferred reader improves average QA from 32.1 for the
no-memory baseline to 38.5.
More importantly, it clearly outperforms the parameter-matched FFN control,
which reaches only 34.5 average accuracy.
The difference cannot be attributed to intrinsic language-model fit:
the FFN-only control actually obtains the lowest perplexity in the table
(7.4 versus 8.7 for transfer), yet remains 4.0 points behind on average QA.

The task-level results show the same pattern.
The transferred reader outperforms the parameter-matched FFN on NQ, WebQA,
TriviaQA, and HotpotQA, with particularly clear margins on NQ
($30.3$ versus $25.5$), TriviaQA ($69.9$ versus $64.1$), and HotpotQA
($27.6$ versus $23.6$).
The FFN control is slightly stronger only on TruthQA.
Thus, simply attaching an equally parameterized target-side residual module
does not reproduce the downstream behavior of transferred memory.

\paragraph{The reader interface is critical.}
Removing context-dependent gating reduces average QA from 38.5 to 33.7,
a drop of 4.8 points.
Replacing the reader with a single affine stitch similarly reduces the
average to 34.3, a 4.2-point decrease.
These results indicate that transfer quality depends not only on the presence
of a memory table, but also on a sufficiently expressive interface for
selecting, transforming, and routing retrieved information into the target
model.

\paragraph{Addressing integrity and memory structure matter.}
Permuting the memory keys while retaining the same underlying table reduces
average QA from 38.5 to 32.5.
This is the largest degradation among the memory-content controls and shows
that preserving the association between addresses and stored values is
important for downstream reuse.

Replacing the transferred table with random memory also decreases average QA
to 34.6.
Interestingly, random memory achieves the same test PPL as the transferred
condition (8.7), yet performs substantially worse on downstream QA.
This provides another example in which intrinsic perplexity alone does not
predict whether a memory representation will be useful for downstream tasks.

\paragraph{Transfer primarily improves initialization rather than the
asymptotic ceiling.}
Training a fresh target memory from scratch eventually reaches an average QA
score of 38.4, essentially matching the 38.5 obtained by transfer.
We therefore do not interpret transfer as providing a substantially higher
asymptotic performance ceiling.
Instead, together with the scaling results in
\Cref{sec:exp:data_efficiency}, the evidence supports a more specific
interpretation:
the transferred memory provides a structured initialization that can reach
useful target-side performance with substantially less adaptation than
learning an equivalent memory representation from scratch.

This distinction also explains why perplexity and downstream QA do not rank
all controls identically.
For example, the parameter-matched FFN reaches the best intrinsic PPL but
substantially worse QA, while scratch memory eventually matches transferred
QA after sufficient target-side optimization.
Successful transfer therefore depends on the interaction between memory
structure, addressing, and the reader interface, rather than on parameter
count or perplexity reduction alone.

\takeawaybox{
The downstream gains cannot be explained by parameter count alone:
successful transfer depends on meaningful memory addressing and a sufficiently
expressive reader interface, while its main advantage over scratch training is
better target-side data efficiency rather than a higher asymptotic ceiling.
}


\section{Conclusion}
\label{sec:conclusion}

We study whether a learned external memory remains useful after it is detached from the backbone that trained it and attached to a different one.
For Engram-style hashed memory, the answer is yes: frozen memory improves the target across tokenizers, hidden sizes, and architectures, including every cell of the main $3 \times 3$ transfer matrix and both directions of peer transfer.
The main lesson is that portability depends not only on the frozen table but also on the target-side reader: stronger readers nearly close the same-model/cross-model gap on QA, and transferred memory clearly beats permuted-key and FFN-only controls while substantially outperforming random memory on the final QA evaluation.
More broadly, cross-model frozen-memory extraction turns external-memory portability into an evaluation framework: a memory artifact should be judged by whether a new backbone can address it, align to it, and extract useful signal without retraining the stored table.

\section*{Acknowledgments}
The authors wish to acknowledge CSC – IT Center for Science, Finland, for computational resources.
Computational facilities were provided by the UTS eResearch High Performance Computer Cluster.
Shaoxiong Ji gratefully acknowledges the support of Foundation PS through the PS Fellowship.

\bibliography{general.bib}

\clearpage
\appendix
\input{XMemTransfer-appendix}

\end{document}

%% file: XMemTransfer-appendix.tex
\section{Supplementary Formalization and Method Details}\label{app:sec:method_details}
\label{sec:formulation}

This section supplies notation and formal details for \Cref{sec:method}.

\subsection{Notations}
\label{app:sec:notations}

\Cref{tab:notation} summarizes the notation used throughout this paper.

\begin{table}[!htbp]
\caption{Notation used throughout this paper.}
\small
\label{tab:notation}
\centering
\vspace{4pt}
\begin{threeparttable}
\begin{tabular}{c|p{0.78\linewidth}}
\toprule
\textbf{Symbol} & \multicolumn{1}{c}{\textbf{Meaning}} \\
\midrule

$A, B$
& Source model and target model, respectively.
\\

$d_A, d_B$
& Hidden dimensions of source model $A$ and target model $B$, respectively.
\\

$\mathcal{E}_A$
& Frozen Engram memory artifact learned with source model $A$, comprising the tables
$\{E_{n,k}\}$.
\\

$E_{n,k}$
& Memory table associated with N-gram order $n$ and hash head $k$,
where $E_{n,k}\in\mathbb{R}^{M\times d_{\text{head}}}$.
\\

$N_{\max}$
& Maximum N-gram order used for memory addressing, with
$n\in\{2,\ldots,N_{\max}\}$.
\\

$K$
& Number of independent hash heads for each N-gram order.
\\

$H$
& Total number of hash heads,
$H=(N_{\max}-1)K$.
\\

$M$
& Number of rows in each hash-head memory table.
\\

$d_{\text{head}}$
& Dimensionality of the embedding row retrieved from one hash-head table.
\\

$\mathbf{e}_t$
& Concatenated memory vector retrieved at token position $t$.
\\

$d_{\text{mem}}$
& Dimensionality of $\mathbf{e}_t$, where
$d_{\text{mem}}=H d_{\text{head}}$.
\\

$\mathcal{P}$
& Canonicalization function mapping a raw decoded string to its canonical form.
\\

$V, V'$
& Raw vocabulary and canonical vocabulary, respectively.
\\

$\varphi_{n,k}$
& Deterministic hash function for N-gram order $n$ and hash head $k$.
\\

$\mathbf{h}_{t,\ell}$
& Target-backbone hidden state at token position $t$ and layer $\ell$.
\\

$\mathcal{L}$
& Set of target-backbone layers at which memory readers are injected.
\\

$S$
& Number of memory injection sites, $S=|\mathcal{L}|$.
\\

$R$
& Number of reader branches at each injection site.
\\

$\mathbf{W}_{K,\ell,r}^{(B)}$
& Branch-specific key-projection matrix for branch $r$ at target layer $\ell$.
\\

$\mathbf{W}_{V,\ell}^{(B)}$
& Value-projection matrix shared across the $R$ branches at target layer $\ell$.
\\

$\alpha_{t,\ell}^{(r)}$
& Context-aware scalar gate for token $t$, layer $\ell$, and branch $r$.
\\

$\beta_{\ell,r}$
& Learnable scalar gate bias for branch $r$ at injection layer $\ell$.
\\

$\mathcal{W}^{(B)}$
& Complete target-side reader parameter set, including all key and value
projections, normalization parameters, and gate biases.
\\

$b, T$
& Batch size and sequence length, respectively.
\\

\bottomrule
\end{tabular}
\end{threeparttable}
\end{table}
\subsection{Memory Transfer Conditions}
\label{sec:formulation:conditions}

Let model~$A$ be a source model equipped with a trained Engram memory table $\mathcal{E}_A$, and let model~$B$ be a target model with hidden dimension $d_B$, tokenizer $\mathcal{T}_B$, and potentially a different architecture.
We ask whether the frozen memory $\mathcal{E}_A$ can provide extractable external knowledge to~$B$ through a lightweight reader $\mathcal{W}^{(B)}$, even when $d_B \neq d_A$ and $\mathcal{T}_B \neq \mathcal{T}_A$.

We allow lightweight target-side reader training, and we do not assume zero-data or universal transfer across all memory architectures.
Under this scope, successful extraction is meaningful only if three conditions can be met simultaneously:

\begin{itemize}
    \item \textbf{Deterministic addressing}: memory lookup indices depend only on the input token sequence through a fixed canonicalization-and-hashing pipeline.
    They do not depend on the backbone's internal hidden states.
    The \emph{same text} can therefore map to the \emph{same memory entries} across backbones, provided the key pipeline is standardized.
    \item \textbf{Architectural decoupling}: the memory table $\mathcal{E}$ is extracted through learned projection matrices $\mathbf{W}_K$ and $\mathbf{W}_V$.
    These projections are natural reader interfaces because they already mediate between the memory space and the backbone hidden space.
    \item \textbf{Gate-based robustness}: even if some transferred memory entries are imperfectly aligned, the sigmoid gate $\alpha_t$ can softly suppress their contribution. 
    In practice, this requires the reader to learn a sufficiently negative match score and/or bias for unhelpful memory, thereby reducing interference with the hidden state.
\end{itemize}

The remainder of this section turns these conditions into a concrete decomposition.
The \emph{key space} determines which memory slot is selected for a piece of text (\Cref{sec:formulation:keys}); the \emph{value space} determines how the vector stored in that slot is interpreted by the target backbone (\Cref{sec:formulation:values}).
Transfer, therefore, requires both address agreement and a reader that maps the retrieved vector into the target residual stream, with gating available to limit interference when alignment is incomplete (\Cref{sec:formulation:gating}).

\subsection{Key-Space Unification}
\label{sec:formulation:keys}

The first obstacle is address compatibility.
In Engram's original design, the hash key is computed over model-specific canonical token IDs.
For example, if model~$A$ assigns token ID 1823 to ``San'' and model~$B$ assigns ID 47, the resulting hash indices differ even when the underlying text is the same, so the two models would read unrelated memory entries.

To remove this mismatch, we define a \emph{tokenizer-agnostic canonical key pipeline}.
Let $\text{decode}(\cdot)$ denote the mapping from a tokenizer's token sequence to its surface string.
For a shared decoded text prefix, both models first normalize the same surface form before hashing:
\begin{equation}
    \mathcal{P}\!\left(\text{decode}(\mathcal{T}_A(x^{\text{text}}_{1:t_A}))\right)
    =
    \mathcal{P}\!\left(\text{decode}(\mathcal{T}_B(x^{\text{text}}_{1:t_B}))\right),
    \label{eq:canon}
\end{equation}
where $x^{\text{text}}_{1:t_A}$ and $x^{\text{text}}_{1:t_B}$ denote source and target token prefixes that decode to the same text span, and $\mathcal{P}$ applies NFKC normalization, lowercasing, and accent stripping.
In this construction, both models map the same decoded text to the same canonical form before hashing, so they can compute identical lookup indices into the shared memory table.

\smallskip\noindent\textbf{Subword boundary misalignment.} The remaining difficulty is that tokenizers may segment the same word into different subword sequences.
Three mitigations are available, in increasing order of robustness: (i) the original NFKC-based canonicalization when vocabularies overlap heavily; (ii) word-boundary N-grams computed over decoded strings; and (iii) character- or byte-level N-gram hashing.
We adopt strategy~(ii) as a practical middle ground: it substantially reduces tokenizer dependence without moving all the way to byte-level modeling.
Our current experiments cover GPT-NeoX (Pythia), LLaMA, Qwen, and Phi tokenizers; languages or tokenizers without clean word boundaries may require the byte-level alternative.

\subsection{Value-Space Alignment via Reader}
\label{sec:formulation:values}

Key agreement ensures that the target model reads the \emph{same} memory entry as the source model; it does not ensure that the retrieved vector is meaningful in the target residual stream.
Model~$A$'s memory vectors $\mathbf{e}_t \in \mathbb{R}^{d_{\text{mem}}}$ were shaped by gradients flowing through~$A$'s hidden geometry, which model~$B$ does not share.

Research on embedding space alignment~\citep{mikolov2013exploiting, kornblith2019similarity, sanjeev2024embedding} suggests that representational spaces of models trained on similar data are often related by an approximately linear transformation.
The Platonic Representation Hypothesis~\citep{huh2024platonic} offers one interpretation of this phenomenon, and \citet{chen2025stitching} showed empirically that affine maps can transfer linear features across LLMs.
This motivates the smallest plausible bridge between source memory and target backbone:
\begin{equation}
    \mathbf{W}_V^{(B)}: \mathbb{R}^{d_{\text{mem}}} \to \mathbb{R}^{d_B}, \qquad \mathbf{W}_K^{(B)}: \mathbb{R}^{d_{\text{mem}}} \to \mathbb{R}^{d_B}.
    \label{eq:adaptor-projections}
\end{equation}
The memory table $\mathcal{E}_A$ remains \emph{frozen}; only the reader parameters $\mathcal{W}^{(B)}$ are trained.
For typical configurations ($d_{\text{mem}} = 512$, $d_B = 4096$), the reader size is
\begin{equation}
    |\mathcal{W}^{(B)}| = 2 \cdot d_{\text{mem}} \cdot d_B + 2 \cdot d_B \approx 4.2\text{M}.
    \label{eq:adaptor-param-count}
\end{equation}
This is well within the LoRA-scale budget and small relative to multi-billion-parameter backbones.
The linear choice is deliberate: it isolates whether cross-model reuse is already feasible under a minimal alignment assumption before introducing richer readers.

\subsection{Gating as a Robustness Mechanism}
\label{sec:formulation:gating}

The transfer mechanism also needs a fallback when the transferred memory is unhelpful.
The context-aware gate provides exactly such a mechanism.
If a memory entry is poorly aligned, for example because of a hash collision or because the source geometry does not translate cleanly to the target, the key vector $\mathbf{k}_t = \mathbf{W}_K^{(B)} \mathbf{e}_t$ will be geometrically dissimilar to the backbone query $\mathbf{h}_t$.
By learning a sufficiently negative match score or bias in those cases, the reader can push $\alpha_t$ toward zero and suppress the memory contribution.
The model can therefore tolerate a non-trivial amount of noisy or only partially aligned transferred memory without forcing that signal into the residual stream.
This self-suppression is analogous to a soft mask in attention, but it operates on the memory channel itself and requires no explicit confidence model.

\subsection{Comparison with Existing Knowledge-Augmentation Methods and Native Engram}
\label{app:comparison}

Our source memory and target reader are trained with next-token prediction rather than task-specific labels.
Consequently, a single fitted reader can be evaluated across multiple downstream tasks without being retrained for each task.
This does not, however, imply universal downstream improvements.
As shown in RQ5 (\Cref{sec:exp:data_efficiency}), transferred memory provides stable intrinsic perplexity gains, whereas downstream improvements remain task-dependent and depend on the alignment among the memory corpus, target model, reader interface, and evaluation objective.

\Cref{tab:comparison} situates frozen-memory extraction relative to retrieval-based, native-memory, and model-editing approaches.
The main design distinction is that our method retains an explicit external memory while making the stored artifact portable across backbones through reader-only fitting.

\begin{table}[ht]
\centering
\small
\caption{Qualitative comparison of knowledge augmentation approaches.
\cmark\ = fully supported, \pmark\ = partially supported, \xmark\ = not supported.}
\label{tab:comparison}
\vspace{4pt}
\resizebox{\columnwidth}{!}{
\begin{tabular}{lccccccc}
\toprule
\textbf{Property} & \textbf{KNN-LM} & \textbf{RAG} & \textbf{RETRO} & \textbf{Mem.\ Layers} & \textbf{LLM Mod.} & \textbf{Engram} & \textbf{Ours} \\
\midrule
$O(1)$ retrieval & \xmark & \xmark & \xmark & \xmark & \xmark & \cmark & \cmark \\
Parametric (trained) & \xmark & \xmark & \cmark & \cmark & \cmark & \cmark & \cmark \\
Cross-model portable & \xmark & \cmark & \xmark & \xmark & \pmark & \xmark & \cmark \\
Surgical deletion & \xmark & \cmark & \xmark & \xmark & \xmark & \cmark & \cmark \\
No context overhead & \xmark & \xmark & \pmark & \cmark & \cmark & \cmark & \cmark \\
Reader-only integration & N/A & N/A & \xmark & \xmark & \xmark & \xmark & \cmark \\
\bottomrule
\end{tabular}%
}
\end{table}

\paragraph{Differences from Native Engram}

\Cref{tab:engram_formula_comparison} makes the architectural changes explicit at the formula level.
Relative to native Engram, the transfer-oriented version uses shared canonical hashing, freezes the exported table, and replaces within-block integration with a target-side residual reader.
\begin{table}[ht]
\centering
\small
\renewcommand{\arraystretch}{1.25}
\setlength{\tabcolsep}{5pt}

\caption{Formula-level comparison between original Engram and our transfer-oriented reader.
We replace the native within-block module with a target-side reader over frozen hashed memory.}
\label{tab:engram_formula_comparison}

\resizebox{\columnwidth}{!}{
\begin{tabular}{p{3.6cm} p{6.8cm} p{6.8cm}}
\toprule
\textbf{Aspect} & \textbf{Original Engram} & \textbf{Transfer-oriented Engram Reader} \\
\midrule

Memory indexing
&
Layer-specific compressed-token $n$-gram hashing: \[ r_t^{(l,n,j)} = \Bigl(
\bigoplus_{k=0}^{n-1} c(x_{t-k}) \cdot a_k^{(l)}
\Bigr) \bmod p_{l,n,j} \]
&
Shared canonical hashing: \[ r_t^{(n,j)} = \bigl( H(\mathrm{canon\_ngram}_t^{(n)}) \oplus s_j \bigr) \bmod M \]
\\

\midrule

Memory representation
&
Layer-specific Engram embedding: \[ e_t^{(l)} = \mathrm{Concat}_{n,j}\mathbf{E}_{l,n,j}[r_t^{(l,n,j)}] \]
&
Shared memory vector: \[
m_t =
\mathrm{Concat}_{j}\mathbf{T}_{j}[r_t^{(j)}] \quad (d_{\mathrm{mem}} = 512) \]
\\

\midrule

Key / gate
&
\begin{equation}
\begin{aligned}
s_t &=
\frac{
\left\langle
\mathrm{RMSNorm}(W_k e_t), \mathrm{RMSNorm}(h_t)
\right\rangle
}{\sqrt{d}} \\
\alpha_t &= \sigma\!\left( \mathrm{sign}(s_t)\sqrt{|s_t|}
\right)
\end{aligned}\notag
\end{equation}
&
\[ k_t^{(r)} = W_k^{(r)} m_t,\quad v_t = W_v m_t \] \[ \alpha_t^{(r)} = \sigma\!\left(
\frac{
\langle \mathrm{RMSNorm}(h_t), \mathrm{RMSNorm}(k_t^{(r)}) \rangle}{\sqrt{d}} + \beta_r
\right)
\]
\\

\midrule

Output update
&
\[
u_t = \alpha_t W_v e_t,\quad
o_t = u_t + \mathrm{ShortConv}(u_t)
\]
&
\[
o_t =
\frac{1}{R}\sum_{r=1}^{R}\alpha_t^{(r)} v_t,
\quad h_t \leftarrow h_t + o_t \]
\\

\midrule

Branch structure
&
Native multi-branch inside backbone
&
Explicit reader branches ($R$ controls capacity)
\\

\midrule

Injection
&
Internal transformer block component
&
Post-layer hook injection (e.g., layers 2 and 10)
\\

\midrule

Training
&
Joint training with backbone
&
Two-stage: source training + frozen memory + target reader fitting
\\

\bottomrule
\end{tabular}
}
\end{table}
\section{Implementation Details}\label{app:sec:implementation_details}
\label{app:impl}

\subsection{Models}
\label{app:models}
We evaluate transfer across six independently developed model families with different architectures, tokenizers, and parameter scales: Pythia~\citep{biderman2023pythia}, TinyLlama~\citep{zhang2024tinyllama}, Qwen3.5~\citep{qwen2025qwen3}, Phi-4-mini~\citep{abouelenin2025phi}, LLaMA~2~\citep{touvron2023llama}, and Mistral~\citep{jiang2023mistral7b}.
These choices are designed to separate several sources of transfer difficulty, including model scale, architectural mismatch, tokenizer mismatch, and differences in pre-training or post-training.

The main transfer matrix uses Pythia, Qwen3.5, and TinyLlama because together they provide both same-family and cross-family transfer settings across a wide range of model sizes.
The Pythia pair offers a controlled same-tokenizer comparison, while transfers to TinyLlama and Qwen3.5 introduce a tokenizer and architectural mismatch.
The peer-transfer study adds Phi-4-mini to test whether memory remains portable between similarly scaled but independently trained model families, rather than only from smaller to larger models.

The corpus-matched downstream suite uses Qwen3.5 source and target models, so that source scale, target scale, and corpus alignment can be varied within a single model family while keeping the broader architecture and tokenizer family controlled.
This makes it easier to attribute downstream differences to memory content, target capacity, and reader fitting, rather than to unrelated model-family-level changes.

The QA experiments in~\Cref{tab:QA_mistral_summary} and associated ablations use LLaMA~2 as the memory source and Mistral as the target backbone.
This pair provides a stronger cross-family setting at the 7B scale, with comparable model capacity but independently trained backbones and distinct tokenization and representation spaces.
It therefore offers a demanding test of whether the reader can recover useful factual knowledge from a frozen source memory without relying on parameter scale or within-family compatibility.

\Cref{tab:model_details} lists the checkpoints, experimental roles, hidden dimensions, checkpoint types, and tokenizer families used in these experiments.

\begin{table*}[ht]
\centering
\caption{
Models and checkpoints used in the transfer experiments. ``Source'' denotes memory construction, and ``target'' denotes reader fitting for a frozen memory.
Here, $d$ is the residual hidden size used by the memory interface.
}
\label{tab:model_details}

\vspace{3pt}
\footnotesize
\renewcommand{\arraystretch}{1.12}
\setlength{\tabcolsep}{1pt}

\begin{tabularx}{\textwidth}{
    @{} l l >{\raggedright\arraybackslash}p{4.35cm} c c l >{\raggedright\arraybackslash}X @{}
}
\toprule
\textbf{Family}
& \textbf{Model}
& \textbf{Experimental role}
& \textbf{Params.}
& \textbf{$d$}
& \textbf{Checkpoint}
& \textbf{Hugging Face identifier} \\
\midrule

\multirow{2}{*}{Pythia}
& Pythia-160M
& Source: main matrix
& 160M
& 768
& Base
& {\scriptsize\texttt{EleutherAI/pythia-160m}} \\

& Pythia-410M
& Target: main matrix
& 410M
& 1024
& Base
& {\scriptsize\texttt{EleutherAI/pythia-410m}} \\

\midrule

\multirow{4}{*}{Qwen3.5}
& Qwen3.5-0.8B
& \makecell[l]{Source: main matrix, scaling\\Target: downstream}
& 0.8B
& 1024
& Base
& {\scriptsize\texttt{Qwen/Qwen3.5-0.8B-Base}} \\

& Qwen3.5-2B
& Target: scaling, downstream
& 2B
& 2048
& Base
& {\scriptsize\texttt{Qwen/Qwen3.5-2B-Base}} \\

& Qwen3.5-4B
& \makecell[l]{Source: peer, downstream\\
               Target: main matrix, peer,\\
               scaling, downstream}
& 4B
& 2560
& Base
& {\scriptsize\texttt{Qwen/Qwen3.5-4B-Base}} \\

& Qwen3.5-9B
& \makecell[l]{Source: main matrix, downstream\\
               Target: scaling, downstream}
& 9B
& 4096
& Base
& {\scriptsize\texttt{Qwen/Qwen3.5-9B-Base}} \\

\midrule

TinyLlama
& TinyLlama-1.1B
& Target: main matrix
& 1.1B
& 2048
& \makecell[l]{Chat-/instruction-\\tuned}
& {\scriptsize\texttt{TinyLlama/TinyLlama-1.1B-Chat-v1.0}} \\

\midrule

Phi
& Phi-4-mini-instruct
& Source and target: peer transfer
& 3.8B
& 3072
& Instruction-tuned
& {\scriptsize\texttt{microsoft/Phi-4-mini-instruct}} \\

\midrule

LLaMA 2
& LLaMA-2-7B
& Source: open-domain QA, ablation
& 7B
& 4096
& Base
& {\scriptsize\texttt{meta-llama/Llama-2-7b-hf}} \\

Mistral
& Mistral-7B-v0.3
& \makecell[l]{Target: open-domain QA, ablation\\
               Source: self-transfer control}
& 7B
& 4096
& Base
& {\scriptsize\texttt{mistralai/Mistral-7B-v0.3}} \\

\bottomrule
\end{tabularx}
\end{table*}
Pythia uses the GPT-NeoX tokenizer, and the Qwen3.5 models share the Qwen3.5 tokenizer.
TinyLlama and LLaMA-2 use LLaMA-family tokenizers, Phi-4-mini-instruct uses the tokenizer distributed with the Phi checkpoint, and Mistral-7B-v0.3 uses the Mistral tokenizer.
These differences in the tokenizer and hidden size require the target reader to bridge representational and, in cross-tokenizer settings, tokenization boundaries rather than simply reusing a dimension-compatible memory interface.

\subsection{Architecture Configuration}
\label{app:architecture_config}

\paragraph{Memory Configuration}
The source Engram memory uses $N_{\max} = 3$ (bigram and trigram), $K = 4$ hash heads per order, table size $M = 65{,}536$, and head dimension $d_{\text{head}} = 64$, yielding $d_{\text{mem}} = 512$ and approximately 33.5M total memory parameters.

\paragraph{Reader Configuration}
We use the reader configuration appropriate to each experimental regime rather than fitting a separate reader for each evaluation benchmark.
For the experiments in \Cref{tab:QA_mistral_summary}, the rows compare a single-layer one-branch reader, a dual-layer one-branch reader, and a dual-layer, four-branch reader.
For the LLaMA-2-7B $\rightarrow$ Mistral-7B-v0.3 transfer ablation experiments in \Cref{tab:ablation}, memory is injected at layers 2 and 10, with four branch-specific key/gating paths at each site and a shared value projection.
This reader is fitted with 2048-token Wikipedia-2021 sequences and is then applied zero-shot to the task prompts, without task-specific reader training.
For the other experiments, we use the minimal single-layer, single-branch reader from \Cref{sec:method:adaptor}.
The reader is attached at the default relative depth $L/3$, where $L$ is the number of transformer layers, and the target backbone and exported memory remain frozen while only the reader is optimized.
The reader is fitted with the default 512-token training sequences and is then applied zero-shot to the task prompts, without task-specific reader training.
These configuration choices reflect the purpose of each experiment.
The QA reader-design study in \Cref{tab:QA_mistral_summary} intentionally varies reader depth and branch capacity to measure how interface expressivity affects extraction from frozen memory, while the ablation study in \Cref{tab:ablation} uses the strongest validated reader to isolate the roles of gating, addressing, and memory content.

\subsection{Data, Training and Evaluation}
\label{app:data}

We select the training corpora to match the purpose of each experimental regime rather than using a single corpus throughout.
WikiText-103 provides a compact and reproducible language-modeling benchmark for the controlled transfer matrix, scaling, and representation analyses.
FineWeb-Edu and the Nemotron-CC subsets provide substantially broader and more domain-diverse text for studying downstream transfer and corpus alignment.
The five QA (NQ, WebQA, TriviaQA, TruthQA, HotpotQA) experiments use Wikipedia-2021  because their evaluation tasks emphasize factual and encyclopedic knowledge.
Finally, LAMBADA, WikiText-103, and C4 are used in the distribution-shift experiments so that each fitted reader can be evaluated both on its training distribution and on corpora with different genres, structures, and provenances.

For \Cref{fig:target_scale_transfer,fig:qwen2b_ppl_scaling,fig:cka} and \Cref{tab:cross_arch,tab:scaling} experiments, we use WikiText-103~\citep{merity2017pointer}.
The loader removes empty records, tokenizes each remaining text with the active model tokenizer without special tokens, concatenates the resulting token stream, and partitions it into fixed, non-overlapping sequences.
The standard sequence length is 512 tokens; the standard train and test splits are used for training and final evaluation, respectively.
Because the loader discards the incomplete final sequence and the evaluation data are batched with ``drop-last'', the test loader contains 34 or 36 batches in the standard runs due to tokenizer-dependent token counts, rather than because the evaluation size is manually set to one of those values.

For experiments in the downstream evaluation and \Cref{tab:domain,tab:corpus_control,tab:mixed_corpus}, we use FineWeb-Edu \citep{lozhkovfineweb} (\texttt{HuggingFaceFW/fineweb-edu}, ``sample-10BT''), a broad educational web corpus, for source-memory and reader fitting.
The domain-alignment experiments additionally use Nemotron-CC~HQ-DQA~\citep{su2025nemotron} and Nemotron-CC~HQ from \texttt{nvidia/Nemotron-CC-v2.1}; the former is a QA-heavy STEM subset and the latter is a broader organic-web subset used as a corpus-size and structure control.
The resulting reader is evaluated zero-shot on RTE~\citep{dagan2005pascal}, BoolQ \citep{clark2019boolq}, openBookQA~\citep{mihaylov2018openbookqa}, SciQ \citep{welbl2017sciq}, TruthfulQA~\citep{lin2022truthfulqa}, and RACE \citep{lai2017race}.

For QA tasks in \Cref{tab:QA_mistral_summary} to \Cref{tab:ablation}, it contains Natural Questions (NQ) \citep{kwiatkowski2019natural}, WebQA~\citep{berant2013semantic}, TriviaQA~\citep{joshi2017triviaqa}, TruthfulQA~\citep{lin2022truthfulqa}, and HotpotQA~\citep{yang2018hotpotqa}.
The default corpus for the LLaMA-2-7B $\rightarrow$ Mistral-7B-v0.3 experiments is the pre-tokenized December-2021 English Wikipedia~\citep{izacard2023atlas} release (\texttt{Rubin-Wei/enwiki-dec2021-preprocessed-mistral}), distributed with the MLP Memory data pipeline~\citep{chen2025mlpmemory}.
It provides train and test splits; the implementation maps a requested validation split to test and, when the active tokenizer differs from Mistral's, decodes and re-tokenizes the windows before chunking them.

The out-of-domain language-modeling evaluation in \Cref{sec:exp:ood} uses LAMBADA \citep{paperno2016lambada}, WikiText-103~\citep{merity2017pointer}, and the English C4 validation set~\citep{raffel2020exploring}.
LAMBADA and WikiText-103 are considered out-of-domain only when the reader is fitted to a different corpus.
LAMBADA contains narrative text requiring long-range contextual prediction, whereas WikiText-103 consists of curated encyclopedic articles; both differ substantially in genre and provenance from broad web corpora such as C4 and FineWeb-Edu.

\smallskip\noindent\textbf{Training protocol.} The principal settings are summarized in \Cref{tab:training_hyperparams}.
All phases optimize the next-token language-modeling objective; no task labels are used to fit a reader.

\begin{table*}[t]
\centering
\caption{
Principal training and evaluation settings. ``LR / warm-up'' reports the learning rate and number of warm-up steps.
Unless otherwise noted, training uses cosine decay, AdamW with weight decay $0.01$, and gradient-norm clipping at $1.0$.
Alternative-corpus experiments follow the intrinsic reader-fitting protocol with LAMBADA or English C4 replacing WikiText-103.
OOD rows are evaluation-only.
}
\label{tab:training_hyperparams}

\vspace{3pt}
\footnotesize
\renewcommand{\arraystretch}{1.16}
\setlength{\tabcolsep}{3.2pt}

\begin{tabularx}{\textwidth}{
@{} >{\raggedright\arraybackslash}p{2.25cm} >{\raggedright\arraybackslash}p{2.25cm} >{\centering\arraybackslash}p{1.65cm} >{\centering\arraybackslash}p{1.65cm} >{\centering\arraybackslash}p{1.65cm} >{\raggedright\arraybackslash}X @{}
}
\toprule
\textbf{Regime / phase}
& \textbf{Corpus}
& \textbf{Budget / cap}
& \textbf{Seq. / batch}
& \textbf{LR / warm-up}
& \textbf{Trainable components and exceptions} \\
\midrule

\multicolumn{6}{c}{\textit{Primary training settings}} \\
\midrule

RQ1~\ref{sec:exp:transferability} Phase~1
& WikiText-103
& 50M tokens
& 512 / 16
& \makecell{$3\times10^{-5}$\\1{,}000}
& Source backbone, memory, and source reader are trainable.
For Qwen3.5 sources, the memory learning rate is $10^{-3}$.
\\

RQ1~\ref{sec:exp:transferability} Phase~2
& WikiText-103
& 20M tokens
& 512 / 16
& \makecell{$3\times10^{-5}$\\500}
& Only the target reader is trainable; the target backbone and transferred memory remain frozen.
\\

Downstream evaluation, Phase~1
& FineWeb-Edu
& 50M tokens
& 512 / 2
& $10^{-3}$ / 1{,}000
& Source backbone, memory, and source reader trained end-to-end; gradient checkpointing \\

Downstream evaluation, Phase~2
& FineWeb-Edu
& 20M tokens
& 512 / 4
& $3\times10^{-5}$ / 500
& Target reader only; target backbone and transferred memory frozen; gradient checkpointing; early stopping patience 5 \\

Table~\ref{tab:QA_mistral_summary} to Table~\ref{tab:ablation} Phase~1
& Wikipedia-2021
& 10--30M tokens
& 2048 / 1
& \makecell{$3\times10^{-5}$\\1{,}000}
& The source memory and reader are trainable; the source backbone is frozen.
Early stopping uses patience 3.
\\

Table~\ref{tab:QA_mistral_summary} to Table~\ref{tab:ablation} Phase~2
& Wikipedia-2021
& 10--30M tokens
& 2048 / 1
& \makecell{$3\times10^{-5}$\\500}
& Only the target reader is trainable; the target backbone and transferred memory remain frozen.
Early stopping uses patience 3.
\\

\midrule
\multicolumn{6}{c}{\textit{Alternative-corpus training}} \\
\midrule

LAMBADA Phase~1
& LAMBADA train split
& 50M tokens
& 512 / 16
& \makecell{$3\times10^{-5}$\\1{,}000}
& Source backbone, memory, and source reader are trainable.
The finite training split is cycled as needed.
\\

LAMBADA Phase~2
& LAMBADA train split
& 20M tokens
& 512 / 16
& \makecell{$3\times10^{-5}$\\500}
& Only the target reader is trainable; the target backbone and transferred memory remain frozen.
\\

C4 Phase~1
& English C4 train stream
& 50M tokens
& 512 / 16
& \makecell{$3\times10^{-5}$\\1{,}000}
& Source backbone, memory, and source reader are trainable.
\\

C4 Phase~2
& English C4 train stream
& 20M tokens
& 512 / 16
& \makecell{$3\times10^{-5}$\\500}
& Only the target reader is trainable; the target backbone and transferred memory remain frozen.
\\

\midrule
\multicolumn{6}{c}{\textit{Out-of-domain evaluation without additional training}} \\
\midrule

LAMBADA
& \texttt{lambada}, test
& 2M eval tokens
& 512 / 16
& ---
& Evaluation only; source memory, reader, and target backbone remain frozen.
\\

WikiText-103
& \texttt{wikitext-103}, test
& 2M eval tokens
& 512 / 16
& ---
& Evaluation-only distribution-shift probe; no trainable components.
\\

C4
& \texttt{allenai/c4}, English validation
& 5M eval tokens
& 512 / 16
& ---
& Evaluation-only streaming web-text probe; no trainable components.
\\

\bottomrule
\end{tabularx}
\end{table*}
For the standard Pythia source configurations, Phase~1 is end-to-end; the large Qwen and LLaMA source runs freeze the source backbone when required by the memory budget.
After Phase~1, the exported memory table is frozen and attached to the target model.
For Phase~2, only the target-side reader is optimized.

\smallskip\noindent\textbf{Conditions and metrics.} Across transfer experiments, we compare a \emph{Baseline} target with no memory, a \emph{Transferred} condition with a trained frozen source memory, and a \emph{Random memory} condition with an untrained table of the same architecture.
Unless noted otherwise, all conditions use three seeds (42, 137, 2024). For the performance in \Cref{tab:QA_mistral_summary,tab:engram_QA_summary_new}, we use F1 score as the main result, and multiple-choice average and follow the official code and checkpoint to evaluate for MLP memory. For the other downstream tasks, we report the accuracy. 
We report test perplexity (PPL, $\downarrow$) with 95\% bootstrap confidence intervals and use paired $t$-tests on per-batch perplexities.
The experiments in the appendix run on a single NVIDIA H100 NVL GPU.
The experiments in the main text are conducted on the LUMI supercomputer operated by CSC -- IT Center for Science, Finland, using the LUMI-G partition equipped with AMD Instinct MI250X GPUs.
\smallskip\noindent\textbf{Evaluation-set overlap audit.} We compare all 31,807 downstream evaluation questions against the exact Phase-1 and Phase-2 training streams used in the QA experiments.
Natural Questions, WebQuestions, and HotpotQA contain no exact question matches.
The remaining exact matches consist of four duplicated TriviaQA rows (0.013\%) and one generic TruthfulQA question without its gold answer.
Partial 8-gram overlaps predominantly correspond to shared phrasing rather than direct question-answer leakage.
Thus, the reader training streams use no benchmark supervision, although they are not strictly overlap-free.
Since the pretraining corpora of the underlying foundation models are unavailable, we interpret the QA results as controlled comparisons rather than as a contamination-free state-of-the-art claim.

\section{Supplementary Evidence for RQ1} \label{app:sec:transferability}

The main 3$\times$3 matrix in \Cref{sec:exp:transferability} answers the primary question of whether a frozen memory remains useful after being moved to a different backbone.
The experiments below test whether that conclusion survives several plausible alternative explanations.
Specifically, we examine whether the gains depend on shared tokenization, arise only when transferring from a smaller to a larger model, disappear for stronger targets, or can be explained by similarity between source and target representations.
\subsection{Full Cross-Architecture Matrix}
\Cref{tab:cross_arch} reports the complete cross-architecture results for all nine source--target combinations used in the main transfer study in~\Cref{fig:transfer_matrix_heatmap}.
For each source memory, we compare three conditions on each target model: the target backbone without memory augmentation (\emph{Baseline}), the frozen source memory integrated through a fitted target reader (\emph{Transferred}), and a size-matched randomly initialized memory evaluated under the same reader fitting protocol (\emph{Random}).
The transferred-memory condition improves test perplexity in every source--target pair, with relative reductions ranging from $1.6\%$ to $15.7\%$.
The strongest improvement occurs for Qwen3.5-0.8B $\rightarrow$ TinyLlama-1.1B, while the random-memory control generally improves less than the corresponding transferred memory.
In that same cell, random memory already reduces PPL by 13.9\%; therefore, the 15.7\% reduction is not evidence that all of the gain comes from transferred content, but reflects a smaller content-specific margin over the random control.

\begin{table}[ht]
\centering
\small
\renewcommand{\arraystretch}{1.15}
\setlength{\tabcolsep}{3.2pt}
\caption{Full cross-architecture transfer matrix: test PPL ($\downarrow$) for 3 source memories $\times$ 3 target models.
\emph{Random} replaces the learned source memory with a size-matched randomly initialized memory under the same fitting protocol.
Mean $\pm$ std over 3 seeds.
Deltas shown next to the transferred result are relative to the no-memory baseline; negative values indicate better PPL.
A displayed standard deviation of $0.0$ indicates a value below $0.05$ after rounding to one decimal place.
}
\label{tab:cross_arch}
\begin{tabular}{lccc}
\toprule
\textbf{Target} & \textbf{Baseline} & \textbf{Transferred} & \textbf{Random} \\
\midrule
\rowcolor{groupgray}
\multicolumn{4}{c}{\textit{Pythia-160M source}} \\
Pythia-410M & $21.9$ & $\mathbf{21.6 \pm 0.1}${\scriptsize\,\pos{-1.6\%}} & $21.9 \pm 0.0$ \\
Qwen3.5-4B & $10.8$ & $10.1 \pm 0.0${\scriptsize\,\pos{-6.8\%}} & $10.3 \pm 0.0$ \\
TinyLlama-1.1B & $10.6$ & $9.5 \pm 0.0${\scriptsize\,\pos{-10.6\%}} & $10.0 \pm 0.0$ \\
\rowcolor{groupgray}
\multicolumn{4}{c}{\textit{Qwen3.5-0.8B source}} \\
Pythia-410M & $22.9$ & $21.4 \pm 0.0${\scriptsize\,\pos{-6.8\%}} & $22.1 \pm 0.0$ \\
Qwen3.5-4B & $10.5$ & $9.6 \pm 0.0${\scriptsize\,\pos{-8.9\%}} & $10.2 \pm 0.0$ \\
\rowcolor{harmonyblue}
TinyLlama-1.1B & $10.8$ & $\mathbf{9.1 \pm 0.0}${\scriptsize\,\pos{-15.7\%}} & $9.3 \pm 0.0$ \\
\rowcolor{groupgray}
\multicolumn{4}{c}{\textit{Qwen3.5-9B source}} \\
Pythia-410M & $22.6$ & $21.6 \pm 0.0${\scriptsize\,\pos{-4.3\%}} & $22.3 \pm 0.0$ \\
Qwen3.5-4B & $10.5$ & $9.4 \pm 0.0${\scriptsize\,\pos{-10.3\%}} & $10.2 \pm 0.0$ \\
TinyLlama-1.1B & $10.4$ & $9.1 \pm 0.0${\scriptsize\,\pos{-12.2\%}} & $9.3 \pm 0.0$ \\
\bottomrule
\end{tabular}%
\end{table}

\subsection{Supplement Transfer Comparisons Experiments}
\label{app:focused_transfer}
This section supplements the transfer comparison experiments in the same-tokenizer and cross-tokenizer transfer settings and compares against parameter-matched baselines and an external retrieval reference.

\Cref{tab:tier1} isolates the simplest transfer setting, where the source and target share the Pythia family and tokenizer, and asks whether frozen memory still helps when comparing against parameter-matched alternatives.
The result is still favorable to the transferred memory.
It outperforms both random memory and the iso-parameter LoRA baselines, while kNN-LM remains a stronger but much more expensive non-parametric reference because it requires a retrieval datastore and nearest-neighbor search at inference time.

\begin{table}[ht]
\centering
\small
\renewcommand{\arraystretch}{1.15}
\setlength{\tabcolsep}{3.2pt}
\caption{Same-tokenizer transfer (Pythia-160M $\to$ Pythia-410M).
Mean $\pm$ std over 3 seeds.
Deltas in the PPL column are relative to the no-memory baseline; negative values indicate better PPL.
LoRA and Cross-LoRA are iso-parameter baselines (1.03M parameters, same 20M token budget).
KNN-LM uses a 5M-token datastore with $k{=}1024$ retrieval at inference time.}
\label{tab:tier1}
\begin{tabular}{lccc}
\toprule
\textbf{Condition} & \textbf{Test PPL $\downarrow$} & \textbf{95\% CI} & \textbf{Params / Cost} \\
\midrule
Baseline (no memory) & $21.9$ & $[20.6, 23.4]$ & --- \\
\rowcolor{groupgray}
\multicolumn{4}{c}{\textit{Transfer Conditions}} \\
Random memory & $21.9 \pm 0.0${\scriptsize\,\pos{-0.2\%}} & $[20.6, 23.3]$ & 1.05M \\
\rowcolor{harmonyblue}
\textbf{Transferred memory} & $\mathbf{21.6 \pm 0.1}${\scriptsize\,\pos{-1.6\%}} & $\mathbf{[20.2, 23.0]}$ & \textbf{1.05M} \\
\rowcolor{groupgray}
\multicolumn{4}{c}{\textit{Parameter-Matched Baselines}} \\
LoRA (rank 7)~\citep{hu2022lora} & $23.3 \pm 0.8${\scriptsize\,\negc{+6.5\%}} & $[21.9, 24.9]$ & 1.03M \\
Cross-LoRA~\citep{xia2025crosslora} & $23.9 \pm 0.6${\scriptsize\,\negc{+9.3\%}} & $[22.5, 25.5]$ & 1.03M \\
\rowcolor{groupgray}
\multicolumn{4}{c}{\textit{External Retrieval Reference}} \\
KNN-LM~\citep{khandelwal2021generalization}$^{\dagger}$ & $20.4${\scriptsize\,\pos{-7.1\%}} & $[19.3, 21.5]$ & 5M-tok store \\
\bottomrule
\multicolumn{4}{l}{\scriptsize $^{\dagger}$Non-parametric; requires $k{=}1024$ nearest-neighbor search per token at inference.}
\end{tabular}%
\end{table}

\Cref{tab:tier2} moves to the harder Pythia $\rightarrow$ TinyLlama case, where tokenizer mismatch is unavoidable, and transfer depends on the shared canonicalization pipeline.
Even in this setting, transferred memory reduces PPL from 10.63 to 9.50 and clearly beats random memory.
This supports the claim that the exported memory is not tied to the source tokenizer as long as the target reader is given a compatible canonical interface.

\begin{table}[ht]
\centering
\small
\renewcommand{\arraystretch}{1.15}
\setlength{\tabcolsep}{3.2pt}
\caption{Cross-tokenizer transfer (Pythia-160M $\to$ TinyLlama-1.1B).
Mean $\pm$ std over 3 seeds.
Deltas in the PPL column are relative to the no-memory baseline; negative values indicate better PPL.
Reader parameters: 2.10M.}
\label{tab:tier2}
\begin{tabular}{lccc}
\toprule
\textbf{Condition} & \textbf{Test PPL $\downarrow$} & \textbf{95\% CI} & \textbf{Params} \\
\midrule
Baseline (no memory) & $10.6$ & $[10.0, 11.3]$ & --- \\
\rowcolor{groupgray}
\multicolumn{4}{c}{\textit{Transfer Conditions}} \\
Random memory & $10.0 \pm 0.0${\scriptsize\,\pos{-5.9\%}} & $[9.4, 10.7]$ & 2.10M \\
\rowcolor{harmonyblue}
\textbf{Transferred memory} & $\mathbf{9.5 \pm 0.0}${\scriptsize\,\pos{-10.6\%}} & $\mathbf{[8.9, 10.2]}$ & \textbf{2.10M} \\
\bottomrule
\end{tabular}%
\end{table}

\subsection{Peer-to-Peer Transfer}
\label{sec:exp:peer}
This section tests whether transfer remains useful once the source and target are both strong mid-scale models rather than a small source feeding a much larger target.
\Cref{tab:peer} shows the results of peer-to-peer transfer between Phi-4-mini~\citep{abouelenin2025phi} (3.8B) and Qwen3.5-4B.

\begin{table}[ht]
\centering
\small
\renewcommand{\arraystretch}{1.15}
\setlength{\tabcolsep}{3.2pt}
\caption{Peer-to-peer transfer between Phi-4-mini (3.8B) and Qwen3.5-4B.
Source memories trained on WikiText-103 (50M tokens).
Mean $\pm$ std over 3 seeds.
Deltas shown next to the transferred result are relative to the no-memory baseline; negative values indicate better PPL.}
\label{tab:peer}
\begin{tabular}{lccc}
\toprule
\textbf{Direction} & \textbf{Baseline} & \textbf{Transferred} & \textbf{Random} \\
\midrule
\rowcolor{harmonyblue}
Phi-4-mini $\rightarrow$ Qwen3.5-4B & $10.5$ & $\mathbf{9.5 \pm 0.0}${\scriptsize\,\pos{-10.1\%}} & $10.2 \pm 0.0$ \\
Qwen3.5-4B $\rightarrow$ Phi-4-mini & $12.6$ & $11.5 \pm 0.0${\scriptsize\,\pos{-9.2\%}} & $12.3 \pm 0.0$ \\
\bottomrule
\end{tabular}%
\end{table}

The gains remain bidirectional: Phi $\rightarrow$ Qwen improves by 10.1\%, and Qwen $\rightarrow$ Phi improves by 9.2\%.
Random memory is substantially weaker in both directions, even though it uses the same target-side reader architecture and optimization budget.
Thus, the observed gains cannot be attributed solely to the additional parameter capacity introduced by the reader.

\subsection{Target Model Scaling}
\label{sec:exp:target_scaling}
\Cref{tab:qwen_scaling} keeps the source fixed and scales the target from 2B to 9B parameters to check whether the transfer disappears once the target becomes stronger.

\begin{table}[ht]
\centering
\small
\renewcommand{\arraystretch}{1.15}
\setlength{\tabcolsep}{3.2pt}
\caption{Target model scaling: Qwen3.5-0.8B source $\to$ Qwen3.5-\{2B, 4B, 9B\}.
Mean $\pm$ std over 3 seeds.
Deltas shown next to the transferred result are relative to the no-memory baseline; negative values indicate better PPL.
Baselines are eval-only (no reader training).}
\label{tab:qwen_scaling}
\begin{tabular}{lccc}
\toprule
\textbf{Target} & \textbf{Baseline} & \textbf{Transferred} & \textbf{Random} \\
\midrule
\rowcolor{harmonyblue}
Qwen3.5-2B & $13.7$ & $\mathbf{11.7 \pm 0.0}${\scriptsize\,\pos{-14.1\%}} & $12.6 \pm 0.0$ \\
Qwen3.5-4B & $10.5$ & $9.6 \pm 0.0${\scriptsize\,\pos{-8.9\%}} & $10.2 \pm 0.0$ \\
Qwen3.5-9B & $9.3$ & $8.5 \pm 0.0${\scriptsize\,\pos{-8.3\%}} & $9.0 \pm 0.0$ \\
\bottomrule
\end{tabular}%
\end{table}

The gains persist across all three target sizes.
The largest relative improvement appears at 2B, but even the 9B target still benefits, which suggests that stronger backbones reduce but do not eliminate the value of the transferred memory.

\subsection{Analysis of Backbone Representations}
\label{sec:exp:cka}

We use centered kernel alignment (CKA) \citep{kornblith2019similarity} to measure the similarity between hidden representations produced by different target backbones.
Linear CKA compares the geometry of two representation matrices and is invariant to isotropic scaling and orthogonal transformations, making it suitable for comparing models with different hidden dimensions.
We use WikiText-103 for training with sequence length 512.

For each backbone, we extract hidden states from the Transformer block located approximately one third of the way through the network, $\lfloor L/3 \rfloor$, where $L$ is the total number of Transformer blocks.
This corresponds to block 4 of 12 for Pythia-160M, block 8 of 24 for Pythia-410M, and block 7 of 22 for TinyLlama-1.1B.
We process the same WikiText-103 input sequences with each model, mean-pool the hidden states over non-padding tokens to obtain one representation per sequence, and compute pairwise linear CKA between the resulting representation matrices.
We select the one-third-depth layer to compare intermediate representations after substantial contextual processing, while avoiding the stronger model-specific specialization that may occur near the output layer.

\begin{figure}[!htbp]
    \centering
    \includegraphics[width=0.65\columnwidth]{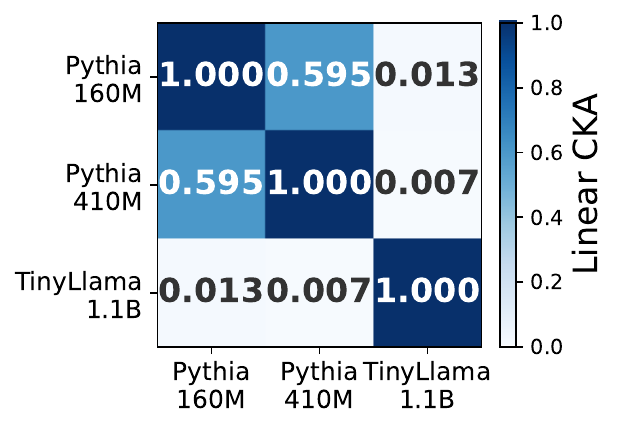}
    \caption{Linear CKA similarity between mean-pooled backbone representations at layer $\lfloor L/3 \rfloor$ for the three models shown in the matrix.}
    \label{fig:cka}
\end{figure}

\Cref{fig:cka} provides a representation-level view of why transfer need not track backbone similarity monotonically.
The two Pythia models exhibit substantially higher representational similarity, with a CKA score of $0.595$, than either model does with TinyLlama, for which the scores are only $0.013$ and $0.007$.
Nevertheless, Pythia-160M $\rightarrow$ TinyLlama-1.1B is among the strongest cross-tokenizer transfer settings in \Cref{tab:cross_arch}, reducing perplexity by $10.6\%$.
By comparison, the more representationally similar Pythia-160M $\rightarrow$ Pythia-410M pair improves perplexity by only $1.6\%$.

Representational similarity alone does not predict the magnitude of transfer gains in these experiments.
Nevertheless, this result is consistent with the reader analysis in \Cref{sec:exp:reader_design}: successful transfer depends on whether the target reader can recover and route useful information from the frozen memory, rather than requiring the source and target backbones to have closely aligned intermediate representation spaces.
This observation complements the main-text finding that reader design, rather than storage capacity alone, determines the utility of a frozen artifact.

\FloatBarrier
\section{Supplementary Analysis of Corpus-Dependent Downstream Transfer for RQ3} \label{app:sec:downstream_boundaries}
These analyses extend RQ3 in \Cref{sec:exp:downstream_boundaries} by separating corpus specialization, reader alignment, mixed-corpus behavior, and out-of-domain effects.

\subsection{Reference Downstream Pattern}
\label{sec:exp:downstream}
The downstream evaluation reports the full six-task evaluation under FineWeb-Edu~\citep{lozhkovfineweb} with aligned Phase~2 training, so the table is not duplicated here.
RTE and SciQ improve most consistently, BoolQ becomes positive for targets of 2B parameters and above, and TruthfulQA remains slightly negative at every scale.
This pattern motivates the following diagnostics: corpus specialization in \Cref{sec:exp:domain}, Phase~2 reader alignment in \Cref{sec:exp:alignment}, mixed-corpus training in \Cref{sec:exp:mixed_corpus}, and unrelated-corpus evaluation in \Cref{sec:exp:ood}.

\subsection{Source-Corpus Specialization}
\label{sec:exp:domain}
\Cref{tab:domain} compares two source-corpus families under matched Phase~2 adaptation: a broad educational web corpus (FW) and a narrower QA-heavy corpus (Nemo HQ-DQA).

\begin{table*}[ht]
\centering
\footnotesize
\renewcommand{\arraystretch}{1.15}
\setlength{\tabcolsep}{2.4pt}
\caption{Domain alignment ablation with corpus-matched Phase~2 readers.
These rows use a separate corpus-alignment control protocol and are not numerically identical to the main FineWeb-Edu values in \Cref{fig:downstream_tasks}.
Positive gains are shown in \textcolor{posgreen}{green} and negative changes in \textcolor{negred}{red}.}
\label{tab:domain}
\begin{tabular}{ll cccccc}
\toprule
\textbf{Target} & \textbf{Source} & \textbf{RTE $\uparrow$} & \textbf{BoolQ $\uparrow$} & \textbf{OpenBookQA $\uparrow$} & \textbf{SciQ $\uparrow$} & \textbf{TruthfulQA $\uparrow$} & \textbf{RACE $\uparrow$} \\
\midrule
\multirow{2}{*}{Qwen3.5-0.8B}
 & FW-avg  & \textbf{\pos{+6.1}} & \negc{-0.6} & \textbf{\pos{+0.7}} & \textbf{\pos{+1.8}} & \negc{-0.1} & \negc{-0.2} \\
 & Nemo-avg & \textbf{\pos{+7.9}} & \textbf{\pos{+4.1}}$^{\ast}$ & \textbf{\pos{+1.1}} & \textbf{\pos{+8.4}}$^{\ast}$ & \negc{-0.8} & \textbf{\pos{+1.1}}$^{\ast}$ \\
\midrule
\multirow{2}{*}{Qwen3.5-2B}
 & FW-avg  & \textbf{\pos{+2.7}} & \textbf{\pos{+3.2}} & \pos{+0.3} & \textbf{\pos{+3.5}} & \negc{-0.4} & \negc{-0.0} \\
 & Nemo-avg & \textbf{\pos{+8.7}}$^{\ast}$ & \textbf{\pos{+14.6}}$^{\ast}$ & \textbf{\pos{+0.5}} & \negc{-0.7} & \pos{+0.0} & \textbf{\pos{+1.1}}$^{\ast}$ \\
\midrule
\multirow{2}{*}{Qwen3.5-4B}
 & FW-avg  & \textbf{\pos{+1.3}} & \negc{-0.2} & \textbf{\pos{+0.6}} & \textbf{\pos{+0.9}} & \negc{-0.5} & \pos{+0.1} \\
 & Nemo-avg & \pos{+0.4} & \textbf{\pos{+0.6}} & \pos{+0.2} & \negc{-6.3} & \negc{-2.9} & \negc{-0.4} \\
\midrule
\multirow{2}{*}{Qwen3.5-9B}
 & FW-avg  & \textbf{\pos{+0.5}} & \textbf{\pos{+0.6}} & \pos{+0.2} & \textbf{\pos{+2.7}} & \negc{-0.7} & \textbf{\pos{+1.0}} \\
 & Nemo-avg & \negc{-4.6} & \textbf{\pos{+2.1}}$^{\ast}$ & \negc{-0.3} & \negc{-3.7} & \pos{+0.2} & \textbf{\pos{+1.4}} \\
\bottomrule
\multicolumn{8}{l}{\scriptsize $^{\ast}$Nemo-avg $>$ FW-avg by $\geq 2$ percentage points.}
\end{tabular}
\end{table*}

The main pattern is that domain-specialized memory helps most on smaller targets and QA-style tasks.
Nemo-avg is especially strong on Qwen3.5-0.8B and Qwen3.5-2B for BoolQ, RTE, and SciQ, but the advantage is less uniform on 4B and 9B targets, showing that specialization helps most when the target still needs strong external support.

\Cref{tab:corpus_control} controls for corpus size on Qwen3.5-2B by comparing an 8B-token STEM question-answering corpus (HQ-DQA) with a broader 26B-token general web corpus (HQ).
Here, the smaller STEM question-answering corpus substantially outperforms the larger general web corpus on BoolQ and RTE and also improves RACE.

\begin{table}[ht]
\centering
\small
\renewcommand{\arraystretch}{1.15}
\setlength{\tabcolsep}{2.4pt}
\caption{Corpus structure vs.\ size on Qwen3.5-2B: accuracy change ($\Delta\%$) under corpus-matched Phase~2 readers.
Positive gains are shown in \textcolor{posgreen}{green} and negative changes in \textcolor{negred}{red}.}
\label{tab:corpus_control}
\begin{tabular}{lcccccc}
\toprule
\textbf{Source corpus} & \textbf{BoolQ $\uparrow$} & \textbf{RTE $\uparrow$} & \textbf{OBQA $\uparrow$} & \textbf{SciQ $\uparrow$} & \textbf{TQA $\uparrow$} & \textbf{RACE $\uparrow$} \\
\midrule
\rowcolor{harmonyblue}
HQ-DQA (8B, STEM Q\&A) & \textbf{\pos{+15.3 $\pm$ 0.3}} & \textbf{\pos{+8.9 $\pm$ 4.7}} & \textbf{\pos{+0.3 $\pm$ 0.2}} & \pos{+0.9 $\pm$ 3.0} & \pos{+0.3 $\pm$ 0.2} & \textbf{\pos{+1.4 $\pm$ 0.4}} \\
HQ (26B, organic web) & \pos{+3.0 $\pm$ 0.4} & \pos{+1.2 $\pm$ 2.1} & \pos{+0.5 $\pm$ 0.1} & \negc{-0.5 $\pm$ 0.0} & \pos{+0.2 $\pm$ 0.3} & \negc{-0.4 $\pm$ 0.2} \\
\rowcolor{groupgray}
FW-avg (reference) & \pos{+3.2} & \pos{+2.7} & \pos{+0.3} & \pos{+3.5} & \negc{-0.4} & \negc{-0.0} \\
\bottomrule
\end{tabular}
\end{table}

\subsection{Phase-2 Reader-Fitting Alignment}
\label{sec:exp:alignment}
\Cref{tab:alignment} probes the failure mode suggested in the main text: stronger source memories can look worse downstream if the target-side reader is trained on a mismatched Phase~2 objective.

\begin{table}[ht]
\centering
\small
\renewcommand{\arraystretch}{1.15}
\setlength{\tabcolsep}{3.2pt}
\caption{Reader alignment: downstream accuracy change ($\Delta\%$) under mismatched vs.\ matched Phase~2 adaptation.
Positive gains are shown in \textcolor{posgreen}{green} and negative changes in \textcolor{negred}{red}.}
\label{tab:alignment}
\begin{tabular}{rlccc}
\toprule
\textbf{Source} & \textbf{Phase 2} & \textbf{BoolQ $\Delta\% \uparrow$} & \textbf{RTE $\Delta\% \uparrow$} & \textbf{SciQ $\Delta\% \uparrow$} \\
\midrule
\rowcolor{groupgray}
\multicolumn{5}{c}{\textit{Mismatched Phase 2 (WikiText-103)}} \\
10M  & WikiText-103 & \pos{+8.3 $\pm$ 0.5} & \pos{+7.7 $\pm$ 3.9} & \negc{-4.5 $\pm$ 0.5} \\
50M  & WikiText-103 & \pos{+6.6 $\pm$ 0.4} & \pos{+6.1 $\pm$ 4.7} & \negc{-1.8 $\pm$ 0.6} \\
200M & WikiText-103 & \pos{+0.0 $\pm$ 0.0} & \pos{+2.9 $\pm$ 4.7} & \negc{-1.4 $\pm$ 0.6} \\
\rowcolor{groupgray}
\multicolumn{5}{c}{\textit{Matched Phase 2 (HQ-DQA)}} \\
10M  & HQ-DQA   & \textbf{\pos{+14.1 $\pm$ 1.2}} & \textbf{\pos{+10.7 $\pm$ 4.0}} & \negc{-2.3 $\pm$ 3.0} \\
\rowcolor{harmonyblue}
50M  & HQ-DQA   & \textbf{\pos{+15.3 $\pm$ 0.3}} & \textbf{\pos{+8.9 $\pm$ 6.2}}  & \pos{+0.9 $\pm$ 3.7} \\
200M & HQ-DQA   & \textbf{\pos{+14.0 $\pm$ 0.8}} & \textbf{\pos{+6.6 $\pm$ 4.9}}  & \textbf{\pos{+2.9 $\pm$ 4.6}} \\
\bottomrule
\end{tabular}
\end{table}

Under WikiText-103 Phase~2 adaptation, increasing source training from 10M to 200M tokens steadily hurts BoolQ transfer.
Once Phase~2 is moved to the matched HQ-DQA distribution, however, all three source budgets recover to roughly 14--15\% BoolQ gain, which shows that the apparent degradation is primarily an alignment problem at the reader rather than a problem with the frozen memory itself.

\Cref{tab:gate_eval} gives a mechanistic view of the same effect through gate statistics.

\begin{table}[!htbp]
\centering
\caption{Gate activation statistics during BoolQ evaluation versus Phase~2 WikiText-103 training.
Longer source training suppresses the gate globally; BoolQ shows a lower mean gate, while the near-closed-gate fraction rises on both distributions under mismatch.}
\label{tab:gate_eval}
\vspace{4pt}
\begin{tabular}{rcccc}
\toprule
\textbf{Source} & \multicolumn{2}{c}{\textbf{Gate mean}} & \multicolumn{2}{c}{\textbf{Frac.\ $< 0.1$}} \\
\cmidrule(lr){2-3} \cmidrule(lr){4-5}
\textbf{tokens} & \textbf{WikiText-103} & \textbf{BoolQ} & \textbf{WikiText-103} & \textbf{BoolQ} \\
\midrule
10M  & 0.6 & 0.5 & 5.6\% & 3.6\% \\
50M  & 0.6 & 0.5 & 9.1\% & 2.9\% \\
200M & 0.5 & 0.4 & 17.5\% & 15.3\% \\
\bottomrule
\end{tabular}%
\end{table}

Longer source training suppresses the gate more strongly overall, and this is visible on BoolQ prompts both in the lower mean gate and in the rise of near-closed gates.
At 200M source tokens, the mean gate on BoolQ drops to 0.352 and the fraction of near-closed gates rises to 15.3\%, while WikiText-103 has a slightly higher near-closed fraction at 17.5\% on the same threshold metric, consistent with broad suppression under reader mismatch rather than a BoolQ-only effect.

\subsection{Broadening Specialist Memory through Corpus Mixing}
\label{sec:exp:mixed_corpus}
\Cref{tab:mixed_corpus} tests whether a strong specialist memory can be broadened without destroying the downstream gains that make it useful in the first place.
The answer is mostly yes.
A 50/50 HQ-DQA + FineWeb-Edu mix preserves the DQA aggregate and slightly exceeds the specialist reference, while sequential mixing and the orthogonal code mixture also retain most of the specialist advantage.
The main tradeoff is that broader mixtures help coverage more than they help the QA-focused aggregate itself.

\begin{table}[ht]
\centering
\small
\renewcommand{\arraystretch}{1.12}
\setlength{\tabcolsep}{1.9pt}
\caption{Mixed-corpus memory study on Qwen3.5-2B.
Positive gains are shown in \textcolor{posgreen}{green} and negative changes in \textcolor{negred}{red}.
The two rightmost columns report aggregate scores: DQA-agg $=$ mean(BoolQ, RTE, SciQ); Br-agg $=$ mean(OBQA, TruthfulQA, RACE).}
\label{tab:mixed_corpus}
\resizebox{\columnwidth}{!}{
\begin{tabular}{lcccccccc}
\toprule
\textbf{Design} & \textbf{BoolQ $\uparrow$} & \textbf{RTE $\uparrow$} & \textbf{OBQA $\uparrow$} & \textbf{SciQ $\uparrow$} & \textbf{TQA $\uparrow$} & \textbf{RACE $\uparrow$} & \textbf{DQA-agg $\uparrow$} & \textbf{Br-agg $\uparrow$} \\
\midrule
\rowcolor{groupgray}
\multicolumn{9}{c}{\textit{Single-corpus references}} \\
HQ-DQA ref. & \textbf{\pos{+15.3 $\pm$ 0.3}} & \textbf{\pos{+8.9 $\pm$ 4.7}} & \pos{+0.3 $\pm$ 0.2} & \pos{+0.9 $\pm$ 3.0} & \pos{+0.3 $\pm$ 0.2} & \pos{+1.4 $\pm$ 0.4} & \textbf{\pos{+8.4}} & \pos{+0.7} \\
FW-Edu ref. & \pos{+2.5 $\pm$ 1.7} & \pos{+1.9 $\pm$ 0.2} & \negc{-0.0 $\pm$ 0.2} & \textbf{\pos{+3.2 $\pm$ 0.6}} & \negc{-0.2 $\pm$ 0.2} & \pos{+0.1 $\pm$ 0.1} & \pos{+2.5} & \negc{-0.1} \\
\rowcolor{groupgray}
\multicolumn{9}{c}{\textit{Mixed or broadened memories}} \\
50/50 mix & \textbf{\pos{+14.3 $\pm$ 1.1}} & \textbf{\pos{+11.7 $\pm$ 3.5}} & \pos{+0.4 $\pm$ 0.4} & \negc{-0.5 $\pm$ 3.8} & \pos{+0.3 $\pm$ 0.3} & \pos{+0.5 $\pm$ 0.3} & \textbf{\pos{+8.5}} & \pos{+0.4} \\
Sequential HQ$\to$FW & \textbf{\pos{+13.6 $\pm$ 1.0}} & \textbf{\pos{+11.9 $\pm$ 2.8}} & \negc{-0.0 $\pm$ 0.2} & \negc{-1.7 $\pm$ 2.7} & \pos{+0.1 $\pm$ 0.4} & \pos{+0.2 $\pm$ 0.3} & \textbf{\pos{+7.9}} & \pos{+0.1} \\
Orthogonal + Code & \textbf{\pos{+14.4 $\pm$ 0.9}} & \textbf{\pos{+9.8 $\pm$ 2.6}} & \pos{+0.4 $\pm$ 0.3} & \negc{-0.4 $\pm$ 3.4} & \negc{-0.2 $\pm$ 0.7} & \pos{+1.3 $\pm$ 0.4} & \textbf{\pos{+7.9}} & \pos{+0.5} \\
\bottomrule
\end{tabular}
}
\end{table}

\subsection{Out-of-Domain Side Effects}
\label{sec:exp:ood}
\Cref{tab:ood} checks whether a reader trained in the WikiText-103 transfer setting systematically distorts language-modeling behavior on unrelated corpora, without additional adaptation.
LAMBADA emphasizes long-range narrative completion, WikiText-103 provides a separately curated language-modeling test set, and C4 represents broad web text; together, they probe distribution shift away from the Phase~2 training corpus.
\begin{table}[ht]
\centering
\small
\renewcommand{\arraystretch}{1.15}
\setlength{\tabcolsep}{3.2pt}
\caption{Out-of-domain evaluation: test PPL ($\downarrow$) for baseline vs.\ transferred memory.
Negative $\Delta$ indicates improvement.}
\label{tab:ood}
\begin{tabular}{llccc}
\toprule
\textbf{Target Model} & \textbf{Dataset} & \textbf{Baseline} & \textbf{Transferred} & \textbf{$\Delta$ (\%)} \\
\midrule
\multirow{3}{*}{Pythia-410M}
& LAMBADA     & $41.0$ & $41.0$ & \negc{+0.0\%} \\
& WikiText-103 & $22.6$ & \textbf{22.1} & \pos{-2.4\%} \\
& C4           & $24.9$ & \textbf{24.8} & \pos{-0.3\%} \\
\midrule
\rowcolor{harmonyblue}
\multirow{3}{*}{TinyLlama-1.1B}
& LAMBADA      & $23.6$ & \textbf{23.5} & \pos{-0.7\%} \\
& WikiText-103 & $10.6$ & \textbf{10.1} & \pos{-5.0\%} \\
& C4           & $11.7$ & \textbf{11.6} & \pos{-0.2\%} \\
\bottomrule
\end{tabular}
\end{table}
The OOD effect is mostly neutral. Both targets get a modest improvement on WikiText-103~\citep{merity2017pointer}, while LAMBADA~\citep{paperno2016lambada} and C4~\citep{raffel2020exploring} stay very close to the baseline in either direction.
This supports the paper's boundary claim that transferred memory is not universally helpful, but it is also usually not catastrophically harmful outside the aligned regime.
\FloatBarrier

\subsection{Reader--Evaluation Distribution Mismatch}
\label{app:cross_corpus_reader}

To isolate whether transfer requires identical reader-training and evaluation distributions, we fix a WikiText-trained Pythia-160M source memory and fit Pythia-410M readers on either WikiText-103 or C4.
We then evaluate each reader on both corpora.

\begin{table}[t]
\centering
\small
\caption{Cross-corpus reader fitting and evaluation.
Lower PPL is better.}
\label{tab:cross_corpus_reader}
\begin{tabular}{llccc}
\toprule
Reader corpus & Eval. corpus & No memory & Transfer & $\Delta$ \\
\midrule
WikiText & WikiText & 22.616 & 22.113 & -0.502 \\
WikiText & C4       & 24.726 & 24.663 & -0.062 \\
C4       & WikiText & 22.616 & 22.538 & -0.078 \\
C4       & C4       & 24.726 & 24.604 & -0.122 \\
\bottomrule
\end{tabular}
\end{table}

Results in \Cref{tab:cross_corpus_reader} show that transfer improves over the no-memory target in all four conditions, so reader-training and evaluation corpora need not be identical. 
However, the substantially larger matched WikiText gain shows that the amount of recoverable signal remains distribution-sensitive.
We therefore distinguish \emph{portability}, which persists under moderate corpus mismatch, from \emph{transfer magnitude}, which depends on alignment among the stored memory, reader-fitting distribution, and evaluation distribution.

\section{Multilingual Transfer}
\label{sec:multi-lingual}
Word-boundary units are the default addressing granularity in our English-like experiments, but they are not appropriate for writing systems without reliable whitespace segmentation.
For such inputs, we replace word units with NFKC-normalized, case-folded non-whitespace Unicode character events and apply the same deterministic hashing procedure to character spans.
This preserves the model-independent addressing principle while changing only the canonical unit used to form $n$-grams.
This character-span variant restores non-degenerate cross-tokenizer addressing for Chinese and Japanese and yields positive transfer gains in both languages, as shown in \Cref{tab:char_span_transfer}.
We then trained Qwen2.5-0.5B source memories and Llama-3.2-1B target readers separately on Chinese and Japanese Wikipedia, using character-span hashing, four branches at layers 2/10, fixed held-out 262K-token test splits, and seeds 42/137/2024.
\begin{table}[t]
\centering
\small
\caption{Character-span addressing for non-segmented languages.
Results are averaged over three seeds.}
\label{tab:char_span_transfer}
\begin{tabular}{lccc}
\toprule
Language & No-memory PPL & Transfer PPL & Rel. improvement \\
\midrule
Chinese  & 20.1248 & $19.6718 \pm 0.0033$ & 2.251\% \\
Japanese & 15.3555 & $15.1985 \pm 0.0251$ & 1.023\% \\
\bottomrule
\end{tabular}
\end{table}

\section{Computational Cost Analysis}\label{app:sec:computational_cost}
\label{app:complexity}

We count multiply-adds (MAdds) for dense projections and report memory addressing separately, since a table lookup is a memory-traffic operation rather than a dense matrix multiplication.
Let $N_{\max}$ denote the maximum n-gram order and let $K$ denote the number of independent hash heads used for each order $n\in\{2,\ldots,N_{\max}\}$.
The total number of hash heads is therefore
\begin{equation}
    H=(N_{\max}-1)K.
\end{equation}
$R$ be the number of reader branches at one injection site, and $S$ be the number of injection sites.
For the default memory configuration, $N_{\max}=3$, $K=4$, $H=8$, and the memory dimension is $d_{\text{mem}}=H d_{\text{head}}=512$, where $d_{\text{head}}$ is the dimension of the embedding row retrieved by one hash head.
Each token therefore requires $H$ deterministic hash indices and $H$ embedding-row reads, together supplying $H d_{\text{head}}=512$ memory values.
This cost is $O(N_{\max}+H)$ hash operations and $O(Hd_{\text{head}})$ memory traffic per site and does not grow with the table size $M$; in particular, it does not perform a nearest-neighbor search over the $M=65{,}536$ rows.

At one injection site of the target-side reader, the shared-value, $R$-branch reader performs one $d_{\text{mem}}\!\times\!d_B$ value projection and $R$ key projections of the same shape, where $d_B$ is the hidden dimension of target model $B$.
The remaining reader operations include RMS normalization, $R$ key-query dot products, sigmoid gates, scalar--vector products, and branch aggregation.
These operations require $O(Rd_B)$ work.
We refer to this as the non-projection, lower-order term because it scales linearly with $d_B$, whereas the dense projections scale as $d_{\text{mem}}d_B$.
With $d_{\text{mem}}=512$, the projection term is substantially larger than the non-projection terms in the evaluated configurations.
Thus, the dominant reader cost for a sequence of length $T$ and batch size $B$ is
\begin{equation}
    C_{\mathrm{reader}} = B T S\left[(R+1)d_{\text{mem}}d_B + O(Rd_B)\right] \quad \text{MAdds},
    \label{eq:reader-compute-cost}
\end{equation}
or, per token and ignoring the lower-order gate terms, $S(R+1)d_{\text{mem}}d_B$.
The single-layer, single-branch reader has $R=S=1$, giving $2\cdot512\cdot d_B$ MAdds per token: $1.05$M at $d_B=1024$ (Pythia-410M) and $4.19$M at $d_B=4096$ (the 7B configurations).
The dual-layer, four-branch QA reader has $S=2$ and $R=4$, so its dominant projection cost is $2\cdot5\cdot512\cdot4096=20.97$M MAdds per token.
The shared value projection is counted once per injection site, not once per branch.

For completeness, the corresponding trainable reader parameter count is
\begin{equation}
    P_{\mathrm{reader}} = S\left[(R+1)d_{\text{mem}}d_B + (R+1)d_B + R\right],
    \label{eq:reader-parameter-cost}
\end{equation}
where the second term accounts for the RMSNorm weights and the final term for the branch-specific gate biases.
This gives approximately $4.20$M parameters for the one-site, one-branch reader at $d_B=4096$, and $21.01$M parameters for the two-site, four-branch reader used in the QA configuration.
These reader parameters are the only additional trainable parameters in Phase~2; the transferred $33.6$M-parameter memory table and the target backbone remain frozen.

The comparison with attention should be read as a projection-only lower bound.
A standard self-attention layer has four dense projections, giving $4d_B^2$ MAdds per token before counting the sequence-dependent attention score and value products.
This is $4.19$M at $d_B=1024$ and $67.11$M at $d_B=4096$, compared with $1.05$M and $4.19$M for the single-branch reader.
The additional attention products contribute $O(Td_B)$ work per token for a length-$T$ sequence, whereas the reader contribution remains constant per token.
Consequently, the reader adds a small fixed overhead to each injected layer and does not change the transformer's quadratic dependence on sequence length.
During Phase~2, freezing the backbone and memory removes their optimizer-state and parameter-update costs, although their forward computation and the backward path needed to train the reader are still executed.
\paragraph{Amortization across consumers.}
The practical advantage of a transferable artifact emerges when the same provider memory is reused across multiple targets.
Using the measured payload times above, the cumulative cost of transfer to $N$ adapted targets is

\[ C_{\mathrm{transfer}}(N) = 0.653 + 3.798N, \]

whereas training a fresh target-specific memory for each consumer costs

\[ C_{\mathrm{fresh}}(N) = 3.871N.
\]

Under these measurements, the one-time provider cost is amortized after approximately nine adapted consumers by payload time.
Using allocated GPU-hours gives a corresponding break-even point of approximately eleven consumers.
These numbers are system- and configuration-specific rather than universal thresholds, but they illustrate the lifecycle regime in which reusable memory artifacts become advantageous.
When the provider reader is directly compatible with the consumer, target-side adaptation is unnecessary: the measured consumer cost is zero training tokens and zero parameter updates, while mean QA rises from 32.12 to 38.31.
The optional reader fitting raises it further to 38.50.
\subsection{Measured System and Lifecycle Cost}
\label{app:measured_cost}
\begin{table}[h]
\centering
\small
\caption{Measured cost for the LLaMA-2-7B$\rightarrow$Mistral-7B-v0.3 using QA experiment configuration on LUMI supercomputer.}
\label{tab:measured_cost}
\resizebox{\linewidth}{!}{
\begin{tabular}{lccccc}
\toprule
Condition & Tokens & Train time & GPU-hours &
Peak inf. memory & Prefill \\
\midrule
No memory
& 0 & 0 & 0
& $13.658\pm0.020$ GiB & $55.10\pm0.06$ ms \\

Phase-1 source artifact
& 4.096M & 0.653 h & 3.154
& -- & -- \\

Fresh Mistral memory
& 19.968M & 3.871 h & 15.760
& -- & -- \\

Matched FFN
& 19.968M & 3.680 h & 14.929
& $13.736\pm0.020$ GiB & $55.32\pm0.49$ ms \\

Transfer, incl.
Phase 1
& 4.096M+19.968M & 4.451 h & 18.623
& $13.881\pm0.020$ GiB & $57.69\pm0.40$ ms \\
\bottomrule
\end{tabular}}
\end{table} 

\Cref{app:measured_cost} shows the cost for transfer experiments from LLaMA-2-7B to Mistral-7B-v0.3. 
Relative to the no-memory target, the transferred configuration adds 0.223 GiB of peak inference memory and 2.60 ms of prefill latency in this measured setup.
The Phase-1 source artifact is a one-time provider cost: the recorded run processes 4.096M tokens in 0.653 hours.
Each adapted consumer then trains only the 21.01M-parameter reader, while the memory and backbone remain frozen.
\section{Ethics and Broader Impact}\label{app:sec:ethics}
\label{app:ethics}
Transferred memory raises both opportunities and risks.
On the positive side, a frozen external memory can reduce repeated backbone retraining and make factual updates more modular.
On the negative side, the same mechanism can propagate memorized or provenance-sensitive content across multiple target models if the source memory is not audited.
Engram's advantage is that its stored knowledge is explicit and addressable: entries can be inspected, removed, or replaced without retraining the backbone.

\section{Limitations}\label{app:sec:limitations}

\noindent\textbf{Scale and availability.} Our experiments cover source and target models up to 9B parameters, but do not systematically characterize how target-side reader capacity should scale with backbone size, memory size, training data, or source--target heterogeneity.
In particular, our two-phase protocol first trains the memory with the source model and then freezes both the transferred memory and the target backbone while adapting only the reader.
We therefore do not explore joint training of the backbone, memory, and reader, nor establish a scaling law relating reader depth, branch count, injection placement, and parameter count to model scale.
Although the present results show that stronger readers substantially improve frozen-memory extraction, it remains unclear whether the same reader configurations remain sufficient for much larger models or whether reader capacity must grow with the backbone or memory.

\smallskip\noindent\textbf{Addressing and language diversity.} Our default word-boundary addressing is effective for the evaluated English-like settings, but it is not universally tokenizer- or language-independent.
In particular, whitespace-based word units degenerate for non-segmented Chinese and Japanese.
We show that replacing word units with normalized character spans repairs this specific failure mode and produces positive transfer on Chinese and Japanese Wikipedia.
These experiments demonstrate that the model-independent addressing principle can extend beyond whitespace-delimited text, but they do not establish a universally optimal canonical unit.
Code, noisy Unicode, mixed-script text, byte-level tokenizers, and additional writing systems may require character-, byte-, or task-specific span definitions.
More generally, portability requires agreement on both the canonicalization procedure and the granularity from which memory addresses are formed.
More aggressive tokenizer mismatch may require a stronger key pipeline than the word-boundary canonicalization used here.

\smallskip\noindent\textbf{Reader scope.} Our capacity diagnostics further show that both interface width and reader placement affect extraction quality; the default $d_{\mathrm{mem}}=512$ setting should therefore be viewed as an evaluated operating point rather than a fixed architectural requirement.
The target-side reader is a central component of our transfer framework, but the present study evaluates only a deliberately lightweight subset of the possible reader design space.
The main transfer matrix uses a single-branch linear reader, while the QA experiments extend it with shared-value multi-branch reading and multi-layer injection.
These controlled designs make it possible to isolate the role of target-side extraction while keeping the number of trainable parameters small.
Nevertheless, more heterogeneous source--target pairs may benefit from nonlinear transformations, adaptive branch aggregation, layer-dependent value projections, or learned injection placement.
Such extensions could improve extraction capacity, but would also increase computation and make it harder to distinguish reusable memory content from capacity introduced by the reader itself.
In addition, our results indicate that the usefulness of context-dependent gating varies with the source and target distributions.
A broader study across corpus shifts and model families is needed to characterize when gating is essential and when a simpler reader is sufficient.